\documentclass[10pt,twocolumn,letterpaper]{article}

\usepackage{cvpr}      % To produce the REVIEW version
\usepackage{amsmath,amsfonts,bm}

\def\eqref#1{equation~\ref{#1}}
\def\1{\bm{1}}

\DeclareMathAlphabet{\mathsfit}{\encodingdefault}{\sfdefault}{m}{sl}
\SetMathAlphabet{\mathsfit}{bold}{\encodingdefault}{\sfdefault}{bx}{n}

\DeclareMathOperator{\Tr}{Tr}

\usepackage{graphicx}
\usepackage{amsmath}
\usepackage{amssymb}
\usepackage{booktabs}

\usepackage{microtype}      % microtypography
\usepackage{graphicx}
\usepackage{multirow}
\usepackage{amsthm}
\usepackage{amsmath,amsfonts,bm}
\usepackage[multiple]{footmisc}

\usepackage{diagbox}
\usepackage{graphbox} %loads graphicx package
\usepackage{enumitem}% http://ctan.org/pkg/enumitem
\usepackage{capt-of}% or \usepackage{caption}
\usepackage{color, colortbl}

\makeatletter
\newcommand*{\rom}[1]{\expandafter\@slowromancap\romannumeral #1@}
\makeatother

\newtheorem{definition}{Definition}[section]

\usepackage{amsmath}
\DeclareMathOperator{\vect}{vec}

\usepackage[pagebackref,breaklinks,colorlinks]{hyperref}

\usepackage[capitalize]{cleveref}
\crefname{section}{Sec.}{Secs.}
\Crefname{section}{Section}{Sections}
\Crefname{table}{Table}{Tables}
\crefname{table}{Tab.}{Tabs.}

\def\cvprPaperID{12459} % *** Enter the CVPR Paper ID here
\def\confName{CVPR}
\def\confYear{2023}

\newcommand{\tb}[3]{\setlength{\tabcolsep}{#2mm}\begin{tabular}{#1}#3\end{tabular}}

\title{\textsc{Kite}: A Kernel-based Improved Transferability Estimation Method}

\author{
\tb{@{}c@{}}{10}{
Yunhui Guo
}\\
\tb{@{}c@{}}{1}{
The University of Texas at Dallas
}}

\begin{document}
\maketitle
%%%%%%%%% ABSTRACT
\begin{abstract}
Transferability estimation has emerged as an important problem in transfer learning. A transferability estimation method takes as inputs a set of pre-trained models and decides which pre-trained model can deliver the best transfer learning performance. Existing methods tackle this problem by analyzing the output of the pre-trained model or by comparing the pre-trained model with a probe model trained on the target dataset. However, neither is sufficient to provide reliable and efficient transferability estimations. In this paper, we present a novel perspective and introduce \textsc{Kite}, as a \underline{K}ernel-based \underline{I}mproved \underline{T}ransferability \underline{E}stimation method. \textsc{Kite} is based on the key observations that the separability of the pre-trained features and the similarity of the pre-trained features to random features are two important factors for estimating transferability. Inspired by kernel methods, \textsc{Kite} adopts \emph{centered kernel alignment} as an effective way to assess feature separability and feature similarity. \textsc{Kite} is easy to interpret, fast to compute, and robust to the target dataset size. We evaluate the performance of \textsc{Kite} on a recently introduced large-scale model selection benchmark. The benchmark contains 8 source dataset, 6 target datasets and 4 architectures with a total of 32 pre-trained models. Extensive results show that \textsc{Kite} outperforms existing methods by a large margin for transferability estimation.
\end{abstract}

%%%%%%%%% BODY TEXT
\section{Introduction}
\label{sec:intro}

Transfer learning has become the standard paradigm for addressing computer vision problems such as recognition \cite{sharif2014cnn,donahue2014decaf}, object detection \cite{girshick2014rich,ren2015faster} and semantic segmentation \cite{long2015fully,badrinarayanan2017segnet}. A generic representation can be transferred to a specific task by fine-tuning the pre-trained model \cite{hinton2007recognize,bengio2012deep}. Compared with training from scratch, transfer learning leads to faster convergence \cite{he2019rethinking,raghu2019transfusion} and better performance \cite{chen2020simple}. The effectiveness of transfer learning motivates researchers to train a large number of pre-trained models \footnote{https://www.tensorflow.org/hub}\footnote{https://pytorch.org/hub/}. Practitioners can readily download these pre-trained models and specialize them for their own tasks. However, making transfer learning more accessible and reliable requires addressing one fundamental question:

\begin{figure}
    \centering
    \includegraphics[width=0.46\textwidth]{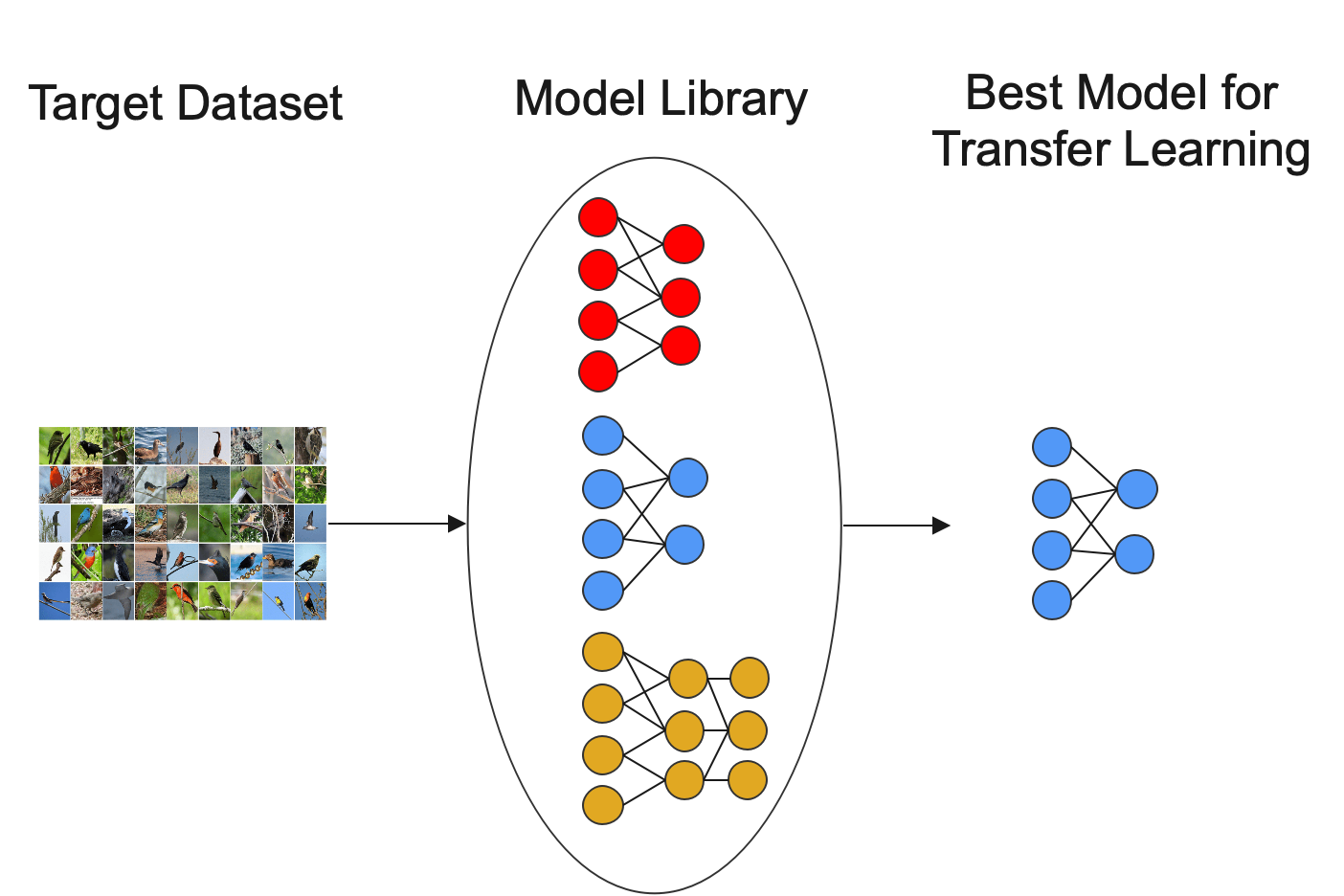}
    \caption{Given a target dataset, the proposed \textsc{Kite} aims to select the best model for transfer learning from a library of pre-trained models.  }
    \label{fig:teaser}
\end{figure}

\begin{center}
    \textit{Given a target dataset, which pre-trained model can deliver \\ the best transfer learning performance?}
\end{center}

There are several factors that make the problem of transferability estimation challenging. First, the number of pre-trained models can be huge. It is thus infeasible to fine-tune each pre-trained model. Second, the pre-trained models are trained with diverse architectures on different source datasets. Third, the target dataset can vary in characteristics and sizes. Although there have been major progresses in addressing these challenges, existing methods rely either on the output of the pre-trained model which is insufficient to provide accurate transferability estimations \cite{tran2019transferability,pandy2022transferability,nguyen2020leep,you2021logme,bao2019information,bolya2021scalable} or on the similarity the pre-trained model with a probe model trained on the target dataset which is time-consuming \cite{dwivedi2019representation,dwivedi2020duality}.

In this paper, we present a novel perspective for transferability estimation by examining the usefulness of pre-trained features from two extremes. At one extreme, the pre-trained features are easy to separate. We find that this is the case if the target task is coarse-grained object classification and the separability of the pre-trained features is a strong indicator for transferability. At the other extreme, the pre-trained features behave similarly to random features. We find that this is case when the target task is fine-grained object classification and the dissimilarity of the pre-trained features to random features serves as a stronger hint. Together, our results show that although these two extremes are related; indeed, better separability generally indicates less similarity to random features. They still have different implications for transferability depending on the target tasks.

Based on the above analysis, we propose an improved transferability estimation method, called \textsc{Kite}, which examines both the separability of pre-trained features and the similarity of the pre-trained features to random features. In particular, \textsc{Kite} leverages \emph{centered kernel alignment} \cite{cortes2012algorithms} which is a standard method for selecting kernels to measure feature separability and feature similarity. As a result, \textsc{Kite} provides an accurate assessment of transfer learning performance without training for a variety of target tasks (See Figure \ref{fig:teaser}). \textsc{Kite} achieves a new state-of-the-art on a large-scale model selection benchmark \cite{bolya2021scalable} which contains a total of 32 pre-trained models. Qualitatively, \textsc{Kite} selects pre-trained models based on the source and target semantics. Quantitatively, \textsc{Kite} produces transferability estimation scores which are well correlated with the final fine-tuning accuracies as measured by Pearson correlation and Weighted Kendall’s $\tau$ rank correlation \cite{vigna2015weighted}. In particular, \textsc{Kite} achieves an improvement of 11.90\% over the state-of-the-art in terms of Pearson correlation.

\noindent \textbf{Contributions.} We highlight the following contributions:

\begin{itemize}[topsep=0pt]
    \item We address transferability estimation from a novel perspective by understanding the effects of feature separability and the similarity of the pre-trained features to random features. This perspective is advantageous since it introduces additional hints for transferability estimation without extra training cost. Our results also shed light on the impact of target tasks on transferability estimation.
    
    \item We propose \textsc{Kite} based on kernel methods for measuring feature separability and feature similarity. We find that feature separability is predictive of transferability only when the pre-trained features are \emph{easy} to separate. When the object categories are hard to separate, the dissimilarity of the pre-trained features to random features serves as a better metric.

    \item We demonstrate that \textsc{Kite} outperforms existing transferability estimation methods on a large-scale model selection benchmark by a large margin in terms of correlation between the final transfer learning accuracies and the transferability estimation scores. \textsc{Kite} is also fast to compute and robust to target dataset variations. 
\end{itemize}

\section{Related Work}

\noindent \textbf{Deep Transfer Learning.} Transfer learning aims at leveraging the knowledge in a pre-trained model to boost the performance on a target dataset \cite{hinton2007recognize,bengio2012deep}. In the standard way of conducting transfer learning, a deep neural network is first pre-trained on a large-scale source dataset such as ImageNet \cite{deng2009imagenet}, then the pre-trained model is fine-tuned on the target dataset with discriminative learning. In particular, the pre-trained model can be trained in a supervised manner with the standard cross-entropy loss \cite{girshick2014rich,ren2015faster,kornblith2019better}. Given unlabeled source datasets, the pre-trained models can also be trained without supervision with contrastive loss \cite{chen2020simple,he2020momentum,grill2020bootstrap}.

Transfer learning has been extensively studied recently due to its theoretical and practical importance. Several works aim at understanding the mechanism behind transfer learning by investigating questions including what is being transferred \cite{neyshabur2020being}, which layer is more transferable \cite{yosinski2014transferable}, what is the correlation between pre-training performance and transfer learning performance \cite{kornblith2019better} and the relation between loss functions and transfer learning performance \cite{kornblith2021better}. Other works focus on developing more sophisticated transfer learning methods with source target joint fine-tuning \cite{ge2017borrowing}, instance-adaptive fine-tuning \cite{guo2019spottune,guo2020adafilter} or additional regularization terms \cite{xuhong2018explicit}.

\noindent \textbf{Transferability Estimation.} Given a target dataset, transferability estimation aims to select the most effective pre-trained model from a library of models for fine-tuning. H-score \cite{bao2019information} proposes to estimate the transferability by seeking pre-trained features with high inter-class variance and low redundancy. NCE \cite{tran2019transferability} and LEEP \cite{nguyen2020leep} estimate transferability by computing the likelihood the target dataset given the pre-trained model. Since likelihood is prone to over-fitting, LogME \cite{you2021logme} proposes to leverage evidence maximization \cite{knuth2015bayesian} for transferability estimation. In particular, LogME computes the logarithm of maximum evidence of the target dataset based on the pre-trained model. More recently, a large-scale transferability estimation benchmark \cite{bolya2021scalable} is proposed for evaluating different methods. PARC is also proposed in \cite{bolya2021scalable} to compute the Spearman correlation between the pre-trained features and target labels for transferability estimation. GBC \cite{pandy2022transferability} is proposed recently to use Bhattacharyya distance to compute class separability for transferability estimation. Methods for Taskonomy Model Selection \cite{zamir2018taskonomy} can also be used for transferability estimation. RSA \cite{dwivedi2019representation} compares the features of the pre-trained model with a ``probe'' model which is trained on the target dataset. In particular, RSA computes the Pearson product-moment correlation coefficient between each pair of images. Spearman correlation is further used to compare the correlation coefficients produced by the pre-trained model and the ``probe'' model. DDS \cite{dwivedi2020duality} generalizes this idea by considering other metrics such as cosine distance and z-score. 

\section{Background}

\noindent \textbf{Transferability Estimation.} We follow the evaluation protocol in \cite{bolya2021scalable} for computing the transferability estimation score and comparing the performance of different transferability estimation methods. Assume that there are $T$ target datasets and $S$ source datasets. For each source dataset $s$, the corresponding pre-trained model is denoted as $M_s$. For each target dataset $t$, we sample $n$ images which lead to a probe set $P_n^t$. Given a transferability estimation method $\mathcal{A}$, the transferability estimation score can be computed based on $P_n^t$ as,
\begin{equation}
    \alpha_{s,t} = \mathcal{A}(M_s, P_n^t)
\end{equation}
Intuitively, $\alpha_{s,t}$ indicates the transferability of the pre-trained model $M_s$ on the target dataset $t$. The ground-truth fine-tuning accuracy of the model $M_s$ on the target dataset $t$ is denoted as $w_{s, t}$. The performance of the transferability estimation method $\mathcal{A}$ is measured via Pearson correlation (PC) \cite{freedman2007statistics} between $ \alpha_{s,t}$ and $w_{s, t}$ for each pre-trained model,
\begin{equation}
    TE(\mathcal{A}) = \frac{1}{T}\sum_{t=1}^T \textnormal{PC}(\{ \alpha_{s,t}: s \in S \}, \{w_{s, t}: s \in S \} )
\end{equation}
%\textcolor{red}{TODO} where $[\alpha_{s,t}]$ and $[w_{s, t}]$ are lists of transferability estimation scores and ground-truth fine-tuning accuracies obtained by each pre-trained model $s$, respectively. 
A larger $TE(\mathcal{A})$ indicates that the transferability estimation score is well correlated with the transfer accuracy, i.e., better transferability estimation. Other metrics such as weighted version of Kendall’s $\tau$ \cite{vigna2015weighted} are also considered in the literature \cite{you2021logme}. 

\noindent \textbf{Kernel Theory.} \textsc{Kite} is rooted in the standard theory of kernels in machine learning \cite{smola1998learning}. Consider a Hilbert space $\mathcal{F}$ which consists of functions from $\mathcal{X}$ to $\mathbb{R}$. $\mathcal{F}$ is a reproducing Kernel Hilbert Space (RKHS) if for each $\mathbf{x} \in \mathcal{X}$, the Dirac evaluation functional $\sigma_\mathbf{x}: \mathcal{F} \rightarrow \mathbb{R}$ is a bounded linear functional. For each RKHS, there exists a unique positive definite kernel $k: \mathcal{X} \times \mathcal{X} \rightarrow \mathbb{R}$ such that for $\mathbf{x} \in \mathcal{X}$ and $\mathbf{x'} \in \mathcal{X}$, there exist corresponding element $\phi(\mathbf{x}) \in \mathcal{F}$ and $\phi(\mathbf{x}') \in \mathcal{F}$ such that $\langle \phi(\mathbf{x}), \phi(\mathbf{x}')  \rangle_\mathcal{F} = k(\mathbf{x}, \mathbf{x'})$. With kernels, instead of computing inner products in the high-dimensional feature space $\mathcal{F}$, we can operate in the input space $\mathcal{X}$. There are several kernel functions such as linear kernel, Gaussian kernel, polynomial kernel and Laplacian kernel \cite{smola1998learning}. In particular, the Gaussian kernel, also called the \emph{Radial Basis Kernel}, is defined as,
\begin{equation}
    k(\mathbf{x}, \mathbf{y}) = \exp ({-\frac{\lVert \mathbf{x} - \mathbf{y} \rVert}{2\sigma^2}})
\end{equation}
where $\sigma$ is a free parameter. 

\noindent \textbf{Centered Kernel Alignment.} 
%Kernel function can be regarded as a way to compare similarity between two samples. 
Given a set of samples $S = \{\bm{x}_1, ..., \bm{x}_n\}$, the kernel matrix $\mathbf{K}$ can be computed by applying the kernel to each pair of samples, i.e., $\mathbf{K}[i, j] = k(\bm{x}_i, \bm{x}_j)$. Different kernel functions give different similarity measures. Kernel alignment was introduced in \cite{cristianini2001kernel} as a means of comparing two kernels. Given two kernels $k_1 : \mathcal{X} \times \mathcal{X} \rightarrow  \mathbb{R}$ and $k_2 : \mathcal{X} \times \mathcal{X} \rightarrow  \mathbb{R}$, we first compute the kernel matrices $\mathbf{K}_1$ and $\mathbf{K}_2$ on the samples $S = \{\bm{x}_1, ..., \bm{x}_n\}$, the alignment of $k_1$ and  $k_2$ is defined as,

\begin{definition}[Alignment \cite{cristianini2001kernel}] 
\begin{equation}
    \rho(\mathbf{K}_1, \mathbf{K}_2) = \frac{\langle \mathbf{K}_1, \mathbf{K}_2 \rangle_F}{ \sqrt{\langle  \mathbf{K}_1, \mathbf{K}_1 \rangle_F \langle  \mathbf{K}_2, \mathbf{K}_2 \rangle_F} }
\end{equation}
where $\langle \rangle_F$ is the Frobenius inner product.

\end{definition}

Kernel alignment has been applied for selecting kernels or combining multiple kernels \cite{cristianini2001kernel}. However, kernel alignment does not correlate well the classification performance \cite{cortes2012algorithms}. This limitation is addressed via centered kernel alignment \cite{cortes2012algorithms} which normalizes the features in the feature space. In particular, the feature mapping is centered by subtracting from the mean as $\phi (\mathbf{x}) - \frac{1}{n}\sum_{i=1}^n \phi(\mathbf{x}_i)$. The corresponding centered kernel matrix $ \mathbf{K}^c$ can be computed from the uncentered kernel matrix $\mathbf{K}$,

\begin{definition}[Centered Kernel Matrix]
\begin{equation}
    \mathbf{K}^c = \left[\mathbf{I} - \frac{\mathbf{1}\mathbf{1}^T}{n} \right] \mathbf{K}  \left[\mathbf{I} - \frac{\mathbf{1}\mathbf{1}^T}{n}\right]
\end{equation}
where $\mathbf{I}$ is the identity matrix and $\mathbf{1} \in \mathbb{R}^{n \times 1}$ is a matrix of ones.

\end{definition}

With the definition of centered kernel matrix, Centered Kernel Alignment (CKA) is defined as follows,

\begin{definition}[Centered Kernel Alignment (CKA) \cite{cortes2012algorithms}]
   \begin{equation}
    \textnormal{CKA}(\mathbf{K}_1^c, \mathbf{K}_2^c) = \frac{\langle \mathbf{K}_1^c, \mathbf{K}_2^c \rangle_F}{ \sqrt{\langle  \mathbf{K}_1^c, \mathbf{K}_1^c \rangle_F \langle  \mathbf{K}_2^c, \mathbf{K}_2^c \rangle_F} }
\end{equation}
\end{definition}

CKA has shown to have a better theoretical guarantee and is better correlated with performance of the kernel on downstream tasks \cite{cortes2012algorithms}.

\section{\textsc{Kite}}
\label{sec:kite}
We begin by discussing the motivation, and then introduce \textsc{Kite} as a kernel-based improved transferability estimation method. Next, we give the interpretation and complexity of \textsc{Kite}. Finally, we compare \textsc{Kite} with existing transferability estimation methods.

\iffalse
\begin{figure*}[t]
   \centering
  \includegraphics[width=0.85\textwidth]{./Figures/all.png}
    \caption{\textbf{The features produced by the most transferable pre-trained model and the random network are drastically different.}. We visualize the features generated by different models on 500 images from CIFAR100 with t-SNE. Pre-trained models which are less transferable produces features that resemble random features. Pre-trained models which are most transferable produces features that roughly capture inter-class variations. }
    \label{fig:motivation} % I can do without the label too
\end{figure*}
\fi

\begin{figure*}[t]
    \centering
    \includegraphics[width=0.73\linewidth]{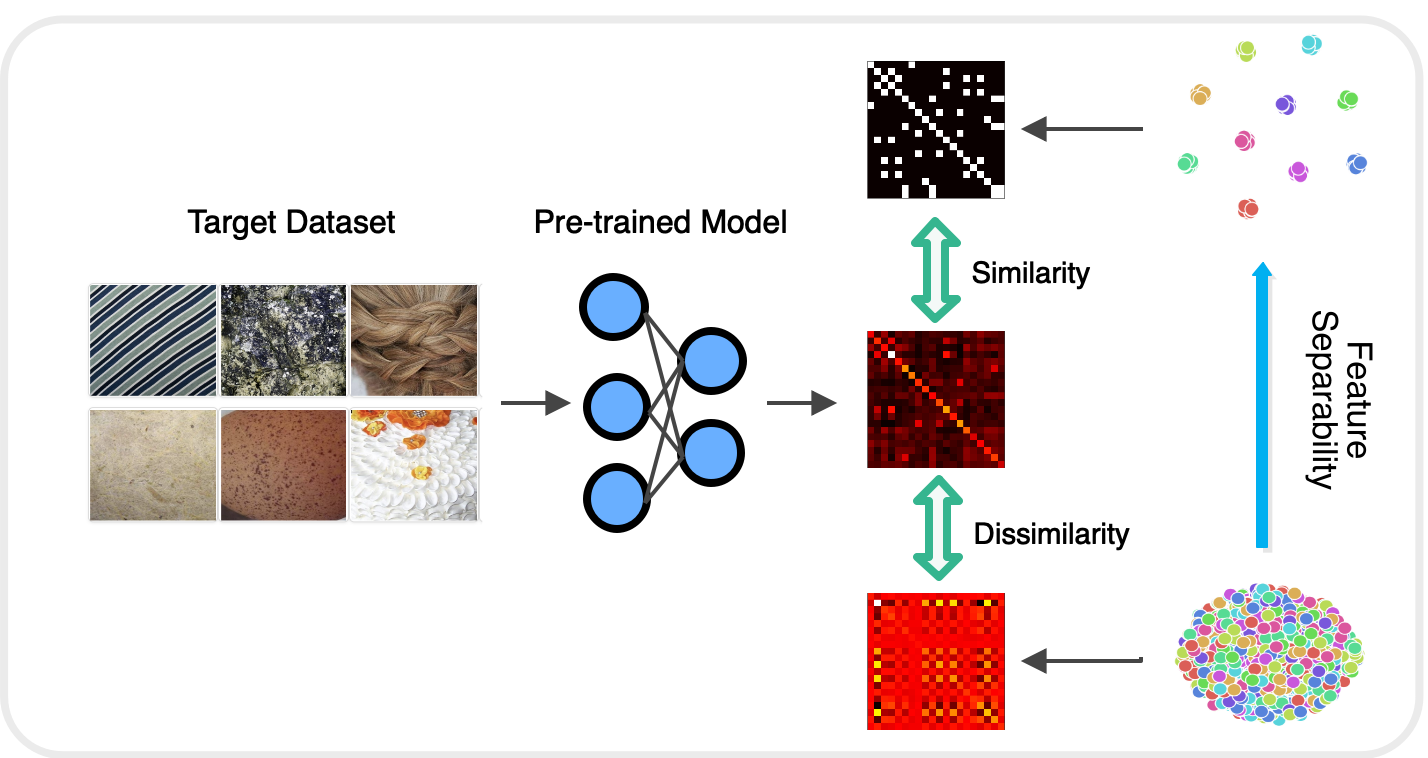}
    \caption{\textbf{\textsc{Kite} considers the separability of pre-trained features and the dissimilarity of the pre-trained features to random features for transferability estimations.} The pre-trained model is first used to generate features for the target dataset. Then, we compute the pre-trained feature kernel matrix, the random feature kernel matrix and the target kernel matrix. CKA is used to compare the (dis)similarity of the pre-trained feature kernel matrix to the random feature kernel matrix and the target kernel matrix.  }
    \label{fig: KITE}
\end{figure*}

\noindent \textbf{Motivation.} 
The usefulness of the pre-trained models naturally depends on the characteristics of the target dataset \cite{neyshabur2020being,kornblith2019better}. For clarity, the pre-trained features are used to refer to the features obtained on the target dataset with the pre-trained model. At one extreme, the pre-trained features are useful and already capture the inter-class variations. At the other extreme, the pre-trained features are similar to random features which fail to capture inter-class separability. Depending on the target task, the usefulness of the pre-trained features can naturally locate anywhere between the two extremes. We find that if the target task is coarse-grained object classification, the pre-trained features are generally easy to separate and the separability is a strong indicator for transferability. However, if the target task is fine-grained object classification, the pre-trained features have poor separability and the dissimilarity of the pre-trained features to random features is a more effective metric for estimating transferability (see Section \ref{sec:effect_data}). 

The above observations lead to the following criteria for estimating transferability which takes the two extreme cases into consideration. A pre-trained model is preferred if  {\bf 1)} it behaves differently from a \emph{random} network and {\bf 2)} it captures inter-class variations. As we will show, better separability indeed means less similarity to random features. However, depending on the target task, the two criteria are still complementary to each other and can be naturally combined for more accurate transferability estimations. The proposed \textsc{Kite} leverages CKA for computing feature separability and feature similarity which can be applied to different target datasets.
\begin{figure*}[!t]
   \centering
    \setlength{\tabcolsep}{0pt}
%\begin{tabular}{@{\hspace{-6pt}}cccc@{}}
\begin{tabular}{cccc}
\includegraphics[width=0.22\textwidth]{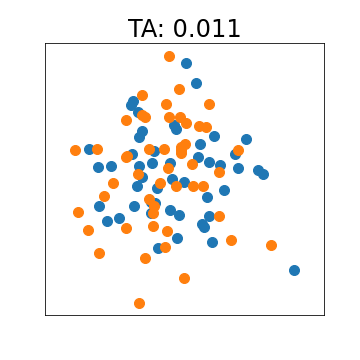} &
\includegraphics[width=0.22\textwidth]{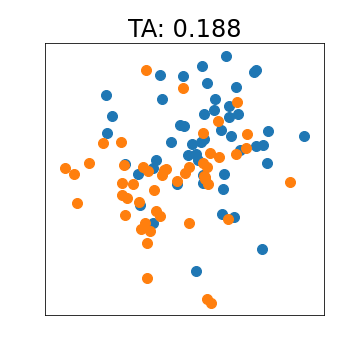} &
\includegraphics[width=0.22\textwidth]{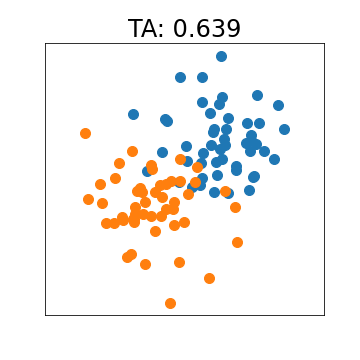} &
\includegraphics[width=0.22\textwidth]{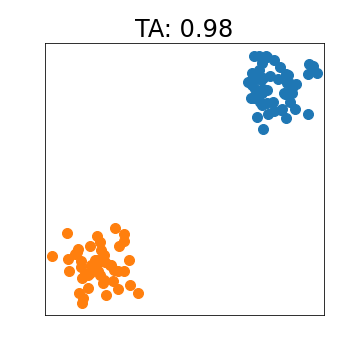} 
\\
\end{tabular}
    \caption{\textbf{TA captures separability of the features}. We validate the effectiveness of TA by generating multiple synthetic datasets. The datasets are generated by sampling from a mixture of two Gaussian distributions with different means. Clearly, the TA score correlates well with feature separability. }
    \label{fig:sepability}
\end{figure*}

\begin{figure*}[!t]
   \centering
 \setlength{\tabcolsep}{0pt}
%\begin{tabular}{@{\hspace{-8pt}}ccc|ccc@{}}
\begin{tabular}{ccc|ccc}
\includegraphics[width=0.16\textwidth]{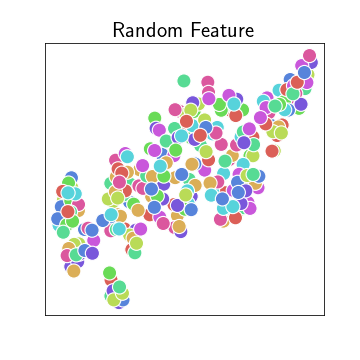}  &
\includegraphics[width=0.16\textwidth]{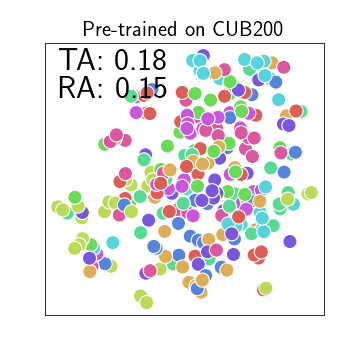}  &
\includegraphics[width=0.16\textwidth]{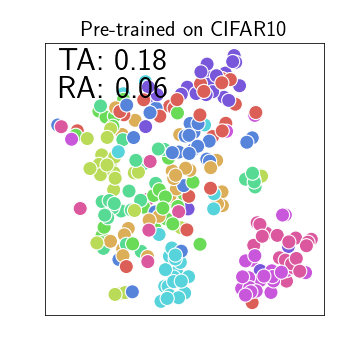}   &
\includegraphics[width=0.16\textwidth]{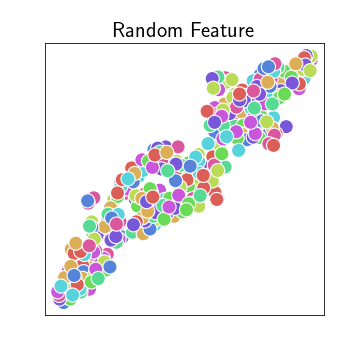}  &
\includegraphics[width=0.16\textwidth]{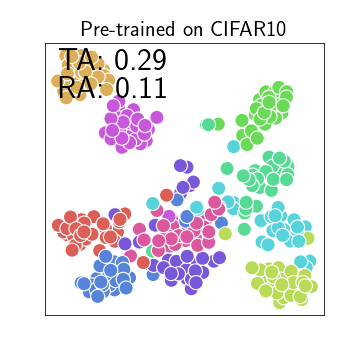} &
\includegraphics[width=0.16\textwidth]{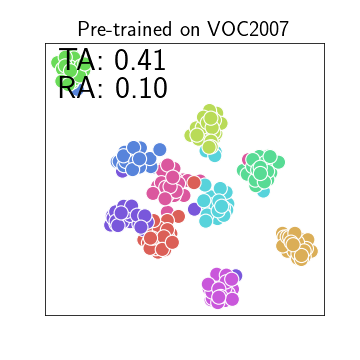} \\
\multicolumn{3}{c}{(a) Target: Stanford Dogs } & \multicolumn{3}{c}{(b) Target: Caltech-101} 
\end{tabular}
    \caption{\textbf{TA and RA uncover different patterns in the feature space.} TA can detect feature separability while RA can expose sample-wise similarity.}
    \label{fig:ra_ta_feature}
\end{figure*}

\noindent \textbf{Computing \textsc{Kite} for Transferability Estimation.} The main idea of \textsc{Kite} is to compute the separability of the pre-trained features and the similarity of the pre-trained features to random features based on CKA. \textsc{Kite} measures how \emph{close} the pre-trained features to the ideal features which perfectly separate the target categories and how \emph{far} the pre-trained features from random features. The random features are generated by using an untrained network on the target dataset. Please see Figure \ref{fig: KITE} for an overview of \textsc{Kite}.

 Given a probe set $P_n^t  = \{(\mathbf{x}_1, y_1), (\mathbf{x}_2, y_2) ..., (\mathbf{x}_n, y_n)\} $ sampled from the target dataset $t$, a model $M_s$ pre-trained on the source dataset $s$ and an untrained random network $M_{random}$. $M_s$ and $M_{random}$ are of the same architecture. Depending on different network initializations, the parameters of $M_{random}$ are different. For simplicity, we assume $M_{random}$ is given which is initialized with some initialization method and the effect of initializations will be investigated in Section \ref{sec:ablation}. We first generate the feature vectors of the probe set using $M_s$ and $M_{random}$ as $F_s  = \{\mathbf{f}_1, \mathbf{f}_2, ..., \mathbf{f}_n\}$ and $F_{random}  = \{ \tilde{\mathbf{f}}_1, \tilde{\mathbf{f}}_2, ..., \tilde{\mathbf{f}}_n\}$, respectively. The feature vector is the output of the model before the classification layer. A kernel function, such as linear kernel or Gaussian kernel, is applied on the features to generate the pre-trained feature kernel matrix $\mathbf{K}_s \in \mathbb{R}^{n \times n}$ and the random feature kernel matrix $\mathbf{K}_{random} \in \mathbb{R}^{n \times n}$. Then the label $y_i$ is converted into one-hot representation. We use $\mathbf{y}_i \in \mathbb{R}^{K \times 1}$ to denote the one-hot encoding of $y_i$, where $K$ is the number of classes. We compute the target kernel matrix $\mathbf{K}_{Y}$ as $\mathbf{K}_{Y}[i, j] = \mathbf{y}_i^T \mathbf{y}_j$. Thus, $\mathbf{K}_{Y}[i, j]$ is 1 if $\mathbf{x}_i$ and $\mathbf{x}_j$ belong to the same class, otherwise $\mathbf{K}_{Y}[i, j]$ = 0. Intuitively, $\mathbf{K}_{Y}$ is the ideal kernel matrix which captures the ground-truth inter-class variations. With the pre-trained feature kernel matrix, the random feature kernel matrix and the target kernel matrix, \textsc{Kite} can be defined as follows,
\begin{definition}[\textsc{Kite}] Given a probe set $P_n^t$ of size $n$ sampled from the target dataset $t$, a pre-trained model $M_s$, an untrained random network $M_{random}$, \textsc{Kite} is defined as,
\begin{equation}
    \textnormal{\textsc{Kite}}(M_s, M_{random}, P_n^t) = \frac{\textnormal{CKA}(\mathbf{K}_s, \mathbf{K}_Y)} { \textnormal{CKA}(\mathbf{K}_{s}, \mathbf{K}_{random})}
\end{equation}
\end{definition}

\noindent \textbf{Remark 1.} The study of the random kernel matrix $\mathbf{K}_{random}$ is prevalent in the literature of random matrix approach to neural networks \cite{pennington2017nonlinear,louart2018random,liao2020random,chen2022concentration}. In particular, it is known that with random Fourier features the random kernel matrix $\mathbf{K}_{random}$ converges to Gaussian kernel matrix as the feature dimension approaches infinity \cite{rahimi2007random}. However, when the activation function is the rectified linear unit $\sigma(x) = \max (0, x)$ and the feature dimension is small, such a limiting behavior is not guaranteed \cite{louart2018random,liao2020random}. In this case, if we vectorize the random kernel matrix as $\vect (\mathbf{K}_{random})$, it can be viewed as a high-dimensional random vector. Moreover, we found that the concentration of $\vect (\mathbf{K}_{random})$ is affected by the target dataset and the architecture. \emph{In the experiments, the random features are computed by averaging the outputs of 5 randomly initialized models.} More discussions can be found in Section D of the Appendix.

\noindent \textbf{Remark 2.} We also consider other ways of combining $\textnormal{CKA}(\mathbf{K}_s, \mathbf{K}_Y)$ and $\textnormal{CKA}(\mathbf{K}_{s}, \mathbf{K}_{random})$. Please see Section \ref{sec:ablation} for  ablation studies.

\noindent \textbf{Interpretation.} We refer to $\textnormal{CKA}(\mathbf{K}_s, \mathbf{K}_Y)$ as \emph{Target Alignment} (TA) which captures feature separability and $\textnormal{CKA}(\mathbf{K}_{s}, \mathbf{K}_{random})$ as \emph{Random Alignment} (RA) which indicates the difference between the pre-trained features and random features. A high \textsc{Kite} score implies that the pre-trained features are different from random features or the pre-trained features are easy to separate. While both Target Alignment and Random Alignment are indicative of transferability, we show that neither of them can decide the transferability of the pre-trained model alone. \textsc{Kite} considers both these two factors which can provide more accurate transferability estimations. Figure \ref{fig:sepability} shows that TA scores correlate well with the separability the features. We create different datasets by sampling from a mixture of two Gaussian distributions with different means and unit variance. By changing the means of the distributions, we can create datasets with different degrees of separability. The TA score increases as the features become more separable. Figure \ref{fig:ra_ta_feature} shows the t-SNE \cite{van2008visualizing} visualizations of the features extracted by different pre-trained models on Stanford Dogs and Caltech-101. It is worth noting that Stanford Dogs is a fine-grained classification task while Caltech-101 is a coarse-grained classification task. It can be observed that the features of Caltech-101 are easier to separate and achieve a higher TA score. This is a general phenomenon for coarse-grained classification tasks as shown in Section B of the Appendix. The behavior of RA is more interesting. On Stanford Dogs, although the features produced by the models pre-trained on CUB200 and CIFAR10 are both hard to separate, the features produced by the model pre-trained on CIFAR10 are far from random: they nearly uncover similarities between the samples from the same class. This is captured via the RA score. In this case, the TA score is not informative since it imposes a much stronger requirement on the feature space. Although TA and RA are generally negatively correlated (see Section B of the Appendix), a small RA score does not necessarily mean a large TA score (see Figure \ref{fig:ra_ta_feature} (a)). 
%We also find that it is generally infeasible to compare the scores across datasets due to the differences of the datasets. 
\textsc{Kite} takes both TA and RA into consideration which can deal with target datasets with different characteristics. Section \ref{sec:effect_data} further shows that TA is particularly effective if the target task is coarse-grained object classification and RA is effective when the target task is fine-grained classification.

\noindent \textbf{Complexity.} \textsc{Kite} only requires forward passes to compute the features of the pre-trained model and the random model, which is inevitable for most of the transferability estimation methods. Given $n$ samples with feature dimension $d$, the complexity of computing the kernel matrices is $\mathcal{O}(n^2d)$. Given that we usually sample a small probe set and the feature dimension is in orders of hundreds or thousands, \textsc{Kite} incurs negligible computational cost. 
%More importantly, \textsc{Kite} does not require any training which makes it well suited for practical applications.

\noindent \textbf{Connection to Existing Methods.} The existing transferability estimation methods roughly fall into two broad categories: {\bf 1)} the estimation is only based on the output the pre-trained model \cite{tran2019transferability,nguyen2020leep,you2021logme,bao2019information,bolya2021scalable,pandy2022transferability} or {\bf 2)} the estimation is based on the comparison of the pre-trained model with a probe model trained on the target dataset \cite{dwivedi2019representation,dwivedi2020duality}. Different from existing methods, \textsc{Kite} takes a novel perspective by considering both feature separability and the similarity of the pre-trained features to random features. As we will show, \textsc{Kite} provides much more accurate transferability estimations while being efficient to compute.

\section{Experiments}
\begin{table*}[t]
    \centering
%\scalebox{0.8}{
    %\begin{tabular}{c|c|c|c|c|c}
\scalebox{0.9}{\begin{tabular}{c<{\ }|c<{\ }|c<{\ }|c<{\ }|r@{ $\pm$ }l<{\ }|r@{ $\pm$ }l<{\ }r@{ $\pm$ }l<{\ }r@{ $\pm$ }l<{\ }}
    \toprule
       \multicolumn{1}{c}{ \textbf{Method}}   & \multicolumn{1}{c}{\textbf{Need Training}} & \multicolumn{1}{c}{\textbf{Input}} & \multicolumn{1}{c}{\textbf{Time (ms)} $\downarrow$}& \multicolumn{2}{c}{\textbf{Mean PC (\%)} $\uparrow$} & \multicolumn{2}{c}{\textbf{Mean $\tau$} $\uparrow$} \\
       \midrule
        NCE \cite{tran2019transferability}  & No & $P_s(\mathbf{x})$, $y$& 12.5  & 2.21 & 0.52  & 0.19 & 0.00 \\
        LEEP  \cite{nguyen2020leep} & No & $P_s(\mathbf{x})$, $y$& 7.8  & 10.83 & 0.13  & 0.20 & 0.00 \\
        LogME \cite{you2021logme} &  No & $M_s(\mathbf{x})$, $y$ & 2139.4& 55.30 & 0.41 &0.47 & 0.00 \\
        \midrule
        H-Score \cite{bao2019information} & No  & $M_s(\mathbf{x})$, $y$ & 83.1& 55.66 & 0.54 & 0.55 & 0.00\\
            RSA \cite{dwivedi2019representation} &  Yes&$M_s(\mathbf{x})$ &222.4 & 5.37 & 0.57 &0.03 & 0.01 \\    
              DDS \cite{dwivedi2020duality} & Yes & $M_s(\mathbf{x})$&188.8  & 10.58 & 0.32 & 0.03 & 0.00  \\
        PARC \cite{bolya2021scalable}  & No & $M_s(\mathbf{x})$, $y$ &149.1 & 59.15 & 1.17 & 0.54 & 0.04 \\
            GBC \cite{pandy2022transferability} & No & $M_s(\mathbf{x})$, $y$ & 432.2  & 51.44& 0.83 & 0.52 &  0.02\\
            \midrule
            
          Logistic \cite{bolya2021scalable} & Yes & $M_s(\mathbf{x})$, $y$ & 338.0 & 58.31 & 2.39 & 0.57 & 0.01 \\
    1-NN CV \cite{bolya2021scalable} & Yes & $M_s(\mathbf{x})$, $y$ & 123.8 & 60.68 & 1.84 & 0.59 & 0.02\\
    5-NN CV \cite{bolya2021scalable} & Yes & $M_s(\mathbf{x})$, $y$ & 138.2&59.73 & 1.75 & 0.58 & 0.01\\
    Heuristic \cite{bolya2021scalable} & No & N/A & N/A & 50.76 & 0.00 & 0.53 & 0.00\\
    \midrule 
\textsc{Kite}  &  No &   $M_s(\mathbf{x})$, $M_{random}(\mathbf{x})$, $y$ &  135.7  &   \textbf{67.90} & \textbf{0.70} & \textbf{0.61} & \textbf{0.02} \\
    \bottomrule
    \end{tabular}}
    \caption{\textbf{\textsc{Kite} improves upon the existing methods for transferability estimation by a large margin both in terms of Pearson Correlation (Mean PC) and weighted Kendall’s $\tau$ (Mean Kendall’s $\tau$).} \textsc{Kite} is also \emph{fast} to compute and requires no training.} 
    \label{tab:results}
\end{table*}

\noindent \textbf{Benchmark.} We adopt the transferability estimation benchmark proposed in \cite{bolya2021scalable} for evaluation. The transferability estimation benchmark consists of 8 source datasets, 6 target datasets and 4 architectures.  The source datasets include ImageNet 1k \cite{deng2009imagenet}, VOC2007 \cite{everingham2010pascal}, Caltech101 \cite{fei2006one}, CIFAR10 \cite{krizhevsky2009learning}, NA Birds \cite{van2015building}, CUB200 \cite{wah2011caltech}, Oxford Pets \cite{parkhi2012cats} and Stanford Dogs \cite{khosla2011novel}. The target datasets include NA Birds, Stanford Dogs, Caltech101, CIFAR10, Oxford Pets, CUB200. The architectures include ResNet-50 \cite{he2016deep}, ResNet-18 \cite{he2016deep}, GoogLeNet \cite{szegedy2015going} and AlexNet \cite{krizhevsky2012imagenet}. There are a total of 32 pre-trained with all the combinations of source datasets and architectures. 
%The ground-truth transfer learning accuracy of each pre-trained model on each target dataset is provided. 
Please refer to \cite{bolya2021scalable} for more details. For each target dataset, we aim to select the most effective pre-trained model from all the pre-trained models for fine-tuning.
%(excluding the pre-trained model trained on the target dataset itself).

\noindent \textbf{Baselines.} We consider the following baselines which include the state-of-the-art methods for transferability estimation: \emph{Probability-based Methods}: NCE \cite{tran2019transferability}, LEEP \cite{nguyen2020leep} and LogME \cite{you2021logme}. \emph{Feature-based Methods}: RSA \cite{dwivedi2019representation}, DDS \cite{dwivedi2020duality}, H-Score \cite{bao2019information}, PARC \cite{bolya2021scalable} and GBC \cite{pandy2022transferability}. \emph{Heuristic-based Methods}: Logistic \cite{bolya2021scalable}, 1-NN CV \cite{bolya2021scalable}, 5-NN CV \cite{bolya2021scalable}, and Heuristic \cite{bolya2021scalable}. In particular, Logistic trains a logistic classifier on 50\% the probe set and uses the accuracy on the other half as the score. K-NN CV (K = 1 or 5) adopts K-nearest neighbors with leave-one out cross-validation. Heuristic simply considers the number of layers in the pre-trained model $\ell_s$, the size of the training dataset $|D_s|$ and target dataset $ |D_t|$,
\begin{equation}
    \textnormal{Heuristic} = \ell_s + \log(|D_s| + |D_t|)
\end{equation}
\noindent \textbf{Metrics.} We consider Pearson Correlation which takes all the pre-trained models into account for comparing the performance as in \cite{bolya2021scalable}. We also consider weighted version of Kendall’s $\tau$ \cite{vigna2015weighted} which focuses more on the top performing models as in \cite{you2021logme}.

\noindent \textbf{Implementation Details.}
We follow the implementation from \cite{bolya2021scalable} for a fair comparison. The input images are resized to 224 $\times$ 224 for all the datasets. The probe set is constructed in a way that there are at least 2 examples for each class. The size of the probe set is 500. The feature dimension is reduced to 32 using principal component analysis (PCA) \cite{abdi2010principal}. The pre-trained models and fine-tuning accuracies are provided by \cite{bolya2021scalable}. For NCE, LEEP, RSA, DDS, Logistics, 1-NN CV, 5-NN CV and Heuristics, we use the implementations from \cite{bolya2021scalable}. For LogME, we use the implementation provided by the original authors \footnote{https://github.com/thuml/LogME}. For GBC, we adopt the official implementation \footnote{https://github.com/google\-research/google\-research/blob/master/stable\_transfer/transferability/gbc.py}. We implement \textsc{Kite} based on the framework provided by \cite{bolya2021scalable} in Python. No other heuristics are added to the methods for a fair comparison. We use linear kernel in the experiments as we will show that \textsc{Kite} is robust to the choices of kernels. The experiments are repeated for 3 runs with different random seeds. The sampled probe set and the initialization of the random network are different across runs. We report both average performance and standard deviation. All the experiments are done on one NVIDIA GeForce GTX GPU.

\begin{table*}[!t]
    \centering
\scalebox{0.9}{\begin{tabular}{c<{\ }|c<{\ }|c<{\ }|c<{\ }|r@{ $\pm$ }l<{\ }|r@{ $\pm$ }l<{\ }r@{ $\pm$ }l<{\ }r@{ $\pm$ }l<{\ }}
    \toprule
       \multicolumn{1}{c}{ \textbf{Method}}   & \multicolumn{1}{c}{\textbf{Need Training}} & \multicolumn{1}{c}{\textbf{Input}} & \multicolumn{1}{c}{\textbf{Time (ms)} $\downarrow$}& \multicolumn{2}{c}{\textbf{Mean PC (\%)} $\uparrow$} & \multicolumn{2}{c}{\textbf{Mean $\tau$} $\uparrow$} \\
RA & No & $M_s(\mathbf{x})$, $M_{random}(\mathbf{x})$ & 83.9 & 56.33 & 0.33 & 0.53 & 0.01 \\
    TA & No &$M_s(\mathbf{x})$, $y$ & 43.5 &   14.87 & 0.21& 0.19 & 0.01  \\
    HSIC & No &  $M_s(\mathbf{x})$, $M_{random}(\mathbf{x})$ &90.9 & -4.04 &1.29 & 0.12 & 0.01 \\
    \midrule 
  \textsc{Kite} &  No &   $M_s(\mathbf{x})$, $M_{random}(\mathbf{x})$, $y$ &   135.7  &   \textbf{67.90} & \textbf{0.70}  & \textbf{0.61} & \textbf{0.02}  \\
    \bottomrule
    \end{tabular}}
    \caption{\textbf{\textsc{Kite} improves upon Random Alignment (RA), Target Alignment (TA) and HSIC by a large margin both in terms of Pearson correlation (Mean PC) and weighted Kendall’s $\tau$ (Mean Kendall’s $\tau$).}}
    \label{tab:alternatives}
\end{table*}

\subsection{The Results of Transferability Estimation}

Table \ref{tab:results} shows the comparison of \textsc{Kite} with all the baselines. Both in terms of Pearson Correlation (\textbf{Mean PC}) and weighted Kendall’s $\tau$ (\textbf{Mean $\tau$}), the proposed \textsc{Kite} improves the state-of-the-art by a large margin. In particular, \textsc{Kite} improves the \textbf{Mean PC} by 11.90\% over 1-NN CV and \textbf{Mean $\tau$} by 3.38\% over 1-NN CV. This shows that \textsc{Kite} can accurately estimate the transferability of the pre-trained model. In terms of computational time, \textsc{Kite} is comparable to other competitive baselines while providing a significant better transferability estimation performance. It is worth noting that 1-NN CV and 5-NN CV does not scale well in terms of the number of samples since they rely on k-nearest neighbors algorithm. In Section \ref{sec:ablation}, we investigate the performance of using Target Alignment and Random Alignment separately. Section \ref{sec:effect_data} shows the effect of target datasets on the performance of \textsc{Kite}. Section \ref{sec:ablation} further shows that \textsc{Kite} is robust to the size of the probe set and feature dimension.

\subsection{Ablation Studies}
\vspace{-0.15cm}
\label{sec:ablation}
\noindent \textbf{\textsc{Kite} vs. other alternatives.} Naturally, two alternatives are TA and RA. We also consider a third alternative called \emph{Hilbert-Schmidt Independence Criterion} (HSIC) which was proposed in \cite{gretton2005measuring} as a measure of dependence between two random variables $X$ and $Y$. Assume that  $\{\mathbf{x}_1, ..., \mathbf{x}_n\}$ and $\{\mathbf{y}_1, ..., \mathbf{y}_n\}$ are drawn from the joint distribution ($X$, $Y$). $\mathbf{K}$ is the kernel matrix computed on $\{\mathbf{x}_1, ..., \mathbf{x}_n\}$ and $\mathbf{L}$ is the kernel matrix computed on $\{\mathbf{y}_1, ..., \mathbf{y}_n\}$. The empirical HSIC can be computed as $\textnormal{HSIC}(X, Y) = \textnormal{HSIC}(\mathbf{K}, \mathbf{L}) =  \frac{1}{(n-1)^2}\Tr(\mathbf{K}\mathbf{H}\mathbf{L}\mathbf{H})$, where $\mathbf{H} = \mathbf{I} - \frac{1}{n}\mathbf{1}\mathbf{1}^T$ is the centering matrix. One idea is to use $\textnormal{HSIC}(\mathbf{K}_s, \mathbf{K}_{random})$ for transferability estimation. Table \ref{tab:alternatives} shows that \textsc{Kite} outperforms all the alternatives by a large margin. There are two main reasons:  {\bf 1)} Different from RA and HSIC, \textsc{Kite} considers target labels to compute feature separability. {\bf 2)} Different from TA, \textsc{Kite} considers the differences of the pre-trained features with random features which can examine if the pre-trained features capture sample-wise similarity.
%Noted that HSIC performs badly since it does not normalize the kernel matrices as in CKA \cite{kornblith2019similarity}.

\noindent \textbf{What is the effect of initialization of the random network?} We consider three commonly used initializations: Xavier normal \cite{glorot2010understanding}, He normal \cite{he2015delving} and He uniform \cite{he2015delving}. The random features are computed by averaging the outputs of 5 randomly initialized models for each choice. Table \ref{tab:init} shows that initializations of the untrained network have little impact on the performance of \textsc{Kite}. Intuitively, initializations have more influence on the learning of the model and no initializations implicitly capture data similarity. We further consider a data-independent and architecture-independent way to generate the random features. In particular, for each dataset and architecture, we sample from a standard normal distribution to generate the random features and the random kernel matrix is computed accordingly. With this choice, \textsc{Kite} achieves a Pearson correlation score of 53.81 $\pm$ 5.36\% which is much lower than using the features generated by the random network. The reason is that by using the random network we can easily capture the scale of the features which is data-dependent and architecture-dependent. Please see Section D of the Appendix for more details.

\iffalse
\begin{table*}[!t]
    \centering
\scalebox{0.8}{
    \begin{tabular}{c|c|c}
    \toprule
        \textbf{Method}  &  \textbf{Fine-Grained Target Datasets}  & \textbf{Coarse-Grained Target Datasets} \\
  RA & 79.83 $\pm$ 0.78  & 32.83 $\pm$ 1.43  \\
TA & -6.02 $\pm$ 0.52  & 50.04 $\pm$ 0.50 \\
\midrule
  \textsc{Kite} & 79.10 $\pm$ 0.89 &  56.71 $\pm$ 0.51 \\
    \bottomrule
    \end{tabular}}
    \caption{\small \textbf{TA is effective when the target task is coarse-grained classification and RA is effective when the target task is fine-grained classification. \textsc{Kite} leverages the advantages of both metrics to provide more accurate transferability estimation.} The Pearson correlations achieved by different methods are shown.  }
    \label{tab:fine_or_coarse}
\end{table*}
\fi

 \begin{table*}[t]
\begin{minipage}[b]{0.45\textwidth}
    \small
     \centering
\scalebox{0.9}{
    \begin{tabular}{c|c|c}
    \toprule
        \textbf{Method}  &  \textbf{Fine-Grained}  & \textbf{Coarse-Grained} \\
  RA & 79.83 $\pm$ 0.78  & 32.83 $\pm$ 1.43  \\
TA & -6.02 $\pm$ 0.52  & 50.04 $\pm$ 0.50 \\
\midrule
  \textsc{Kite} & 79.10 $\pm$ 0.89 &  56.71 $\pm$ 0.51 \\
    \bottomrule
    \end{tabular}}
    \captionof{table}{\small  \textbf{TA is effective when the target task is coarse-grained classification and RA is effective when the target task is fine-grained classification. \textsc{Kite} leverages the advantages of both metrics to provide more accurate transferability estimation.} The Pearson correlations achieved by different methods are shown. }
    \label{tab:fine_or_coarse}
\end{minipage}
\hfill
\begin{minipage}[b]{0.45\textwidth}
\begin{tabular}{c}
   \includegraphics[width=0.75\linewidth]{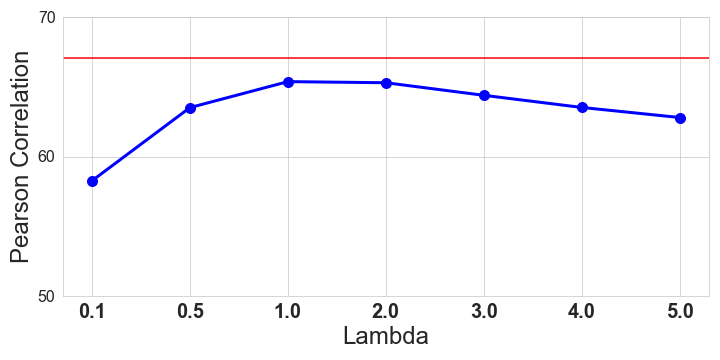} 
\end{tabular}
\captionof{figure}{\small  \textbf{\textsc{Kite} is better than the linear combination alternative}. The \textcolor{red}{red} horizontal line denotes the result of \textsc{Kite}.}
    \label{fig:lambda}
\end{minipage}
\end{table*}

\begin{figure}[!t]
   \centering
    \setlength{\tabcolsep}{0pt}
\begin{tabular}{cc}
   \includegraphics[width=0.45\linewidth]{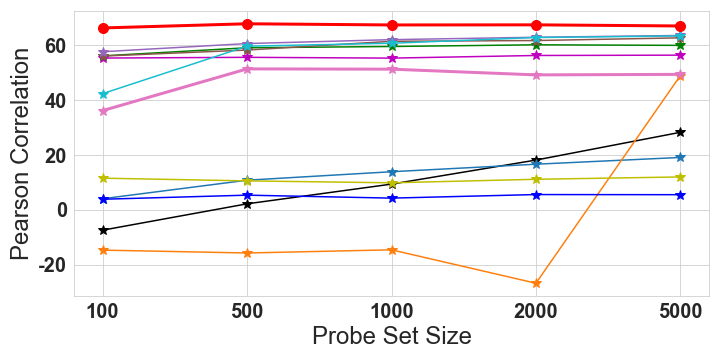} &
   \includegraphics[width=0.45\linewidth]{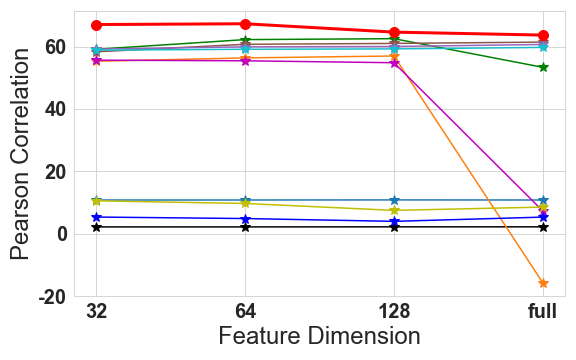} \\
      a) Probe set size & b) Feature dimension  \\
   \multicolumn{2}{c}{\includegraphics[width=0.95\linewidth]{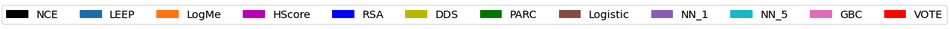}}  \\
\end{tabular}
    \caption{\small \textbf{\textsc{Kite} is robust to the probe set size and feature dimension.} a) The change of Pearson correlation as we change the size of the probe set. b) The change of Pearson correlation as we change the feature dimension.}
    \label{fig:ablation_size}
\end{figure}

\noindent \textbf{Does the size of the probe set matter?} For each target dataset, we vary the probe set size in \{100, 500, 1000, 2000\}. Figure \ref{fig:ablation_size} a) shows that \textsc{Kite} achieves the best results across all the cases. Generally, we observe that across all the methods the performance improves as the probe set size increases. However, the performance does not increase much as the probe set grows even larger. This indicates that more target samples may introduce noises which is challenging for all transferability estimation methods. How to leverage large probe sets for transferability estimation is an interesting future direction.

 \begin{table*}[t]
\begin{minipage}[b]{0.47\textwidth}
    \small
     \centering
   \scalebox{0.9}{ \begin{tabular}{c|c|c}
    \toprule
\textbf{Kernel}   & \textbf{Mean PC (\%) $\uparrow$} & \textbf{Mean $\tau$} $\uparrow$\\
Linear  &67.90 $\pm$ 0.70 &  0.61 $\pm$ 0.02 \\
Laplacian  &  65.51 $\pm$ 0.62  & 0.61 $\pm$ 0.01  \\
Gaussian  &  66.66 $\pm$ 0.47 &  0.62 $\pm$ 0.01 \\
    \bottomrule
    \end{tabular}}
    \captionof{table}{\small  \textbf{\textsc{Kite} is robust to the choices of kernels.} We consider linear, Laplacian and Gaussian kernel for computing the kernel matrices.}
    \label{table:kernels}
\end{minipage}
\hfill
\begin{minipage}[b]{0.47\textwidth}
    \small
     \centering
    \scalebox{0.9}{\begin{tabular}{c|c|c}
    \toprule
\textbf{Init}   & \textbf{Mean PC (\%) $\uparrow$} & \textbf{Mean $\tau$} $\uparrow$\\
 He Normal  &67.90 $\pm$ 0.70  &  0.61 $\pm$ 0.02 \\
 He Uniform  & 66.37 $\pm$ 1.08  &  0.63 $\pm$ 0.01 \\
Xavier Normal & 68.42 $\pm$ 0.56  & 0.62 $\pm$ 0.01 \\
    \bottomrule
    \end{tabular}}  
    \captionof{table}{\small  \textbf{\textsc{Kite} is robust to different initializations of the random network}. We consider Xavier normal, He normal and He uniform.}
    \label{tab:init}
\end{minipage}
\end{table*}

\noindent \textbf{Does the feature dimension matter?} 
We use principal component analysis (PCA) \cite{abdi2010principal} to reduce the feature dimension to 32, 64 and 128. We also consider using the original feature dimension which is denoted as \texttt{full}.  Figure \ref{fig:ablation_size} (b) shows that \textsc{Kite} is robust to the change of feature dimension and achieves the best results across all the cases. 

\noindent \textbf{What are the impacts of different kernel functions?} We investigate the choices the kernel functions on the performance of \textsc{Kite}. We consider linear, Gaussian and Laplacian kernel. Table \ref{table:kernels} shows that \textsc{Kite} is robust to the choices of kernels. This allows \textsc{Kite} to leverage simple linear kernels to evaluate the kernel matrices which is efficient to compute.

\noindent \textbf{Linear combination.} \textsc{Kite} computes the ratio between $\textnormal{CKA}(\mathbf{K}_s, \mathbf{K}_Y)$ and $\textnormal{CKA}(\mathbf{K}_{s}, \mathbf{K}_{random})$. Here we consider linearly combining the two factors as $\textnormal{CKA}(\mathbf{K}_s, \mathbf{K}_Y) - \lambda \textnormal{CKA}(\mathbf{K}_{s}, \mathbf{K}_{random})$, where $\lambda$ is a hyperparameter. Figure \ref{fig:lambda} shows that \textsc{Kite} is better than the linear combination alternative. One plausible reason is that the hyperparameter $\lambda$ is dataset-dependent. By computing the ratio between $\textnormal{CKA}(\mathbf{K}_s, \mathbf{K}_Y)$ and $\textnormal{CKA}(\mathbf{K}_{s}, \mathbf{K}_{random})$ as in \textsc{Kite}, we naturally balance the two factors in a dataset-dependent manner and the effect of $\textnormal{CKA}(\mathbf{K}_{s}, \mathbf{K}_{random})$ is magnified.

\vspace{-0.2cm}
\subsection{The Effect of Target Datasets}
\vspace{-0.1cm}
\label{sec:effect_data}
The effects of target datasets are largely overlooked in the literature of transferability estimation. We find that TA and RA behave rather differently on fine-grained and coarse-grained classification tasks. We use TA to measure and rank the separability of the features. The most separable datasets are three coarse-grained classification tasks: CIFAR10, Oxford Pets and Caltech-101. The least separable datasets are three fine-grained classification tasks: CUB-200, Stanford Dogs and NABirds. Detailed scores are shown in Section B of the Appendix. Table \ref{tab:fine_or_coarse} shows the average Pearson correlations achieved by TA, RA and \textsc{Kite}. First, it can be observed that TA is particularly effective for coarse-grained classification tasks. Second, when the features are hard to separate as in the case of fine-grained classification, RA emerges to be more effective. \textsc{Kite} combines TA and RA which can provide accurate transferability estimations regardless of the characteristics of the target tasks. More discussions can be found in Section B of the Appendix.

\vspace{-0.3cm}
\section{Summary}

We bring a new perspective and propose an effective method called \textsc{Kite} for transferability estimation. \textsc{Kite} estimates transferability by assessing feature separability and comparing the pre-trained model with a random network based on centered kernel alignment. Extensive experiments demonstrate the effectiveness of \textsc{Kite} over the existing transferability estimation methods. We discuss the limitations and future directions in Section E of the Appendix.

\newpage

%%%%%%%%% BODY TEXT - ENTER YOUR RESPONSE BELOW
\section{Appendix}
We propose both a novel perspective and an effective method for transferability estimation. Our core idea has two folds: {\bf 1)} computing the separability of pre-trained features and {\bf 2)} assessing the dissimilarity of the pre-trained features to random features. We demonstrate the state-of-the-art performance for transferability estimation on a large-scale benchmark. In this appendix, we include more details and results:

\begin{itemize}
    \item We show the models selected by \textsc{Kite} for each target dataset in Section \ref{sec:model_selection}.
    \item We show the TA and RA scores for each target dataset and have a detailed analysis in Section \ref{sec:ta_vs_ra}.
    \item We show the results of the methods on each target dataset separately in Section \ref{sec:per_class}.
    \item We add more discussions on the random kernel matrix in Section \ref{sec:random_kernel}.
    \item We discuss the limitations and extensions of \textsc{Kite} in Section \ref{sec:limitation}.
\end{itemize}

\section{Which Pre-trained Model is Selected by \textsc{Kite}?}
\label{sec:model_selection}
\begin{figure}[!h]
    \centering
    \includegraphics[width=0.8\linewidth]{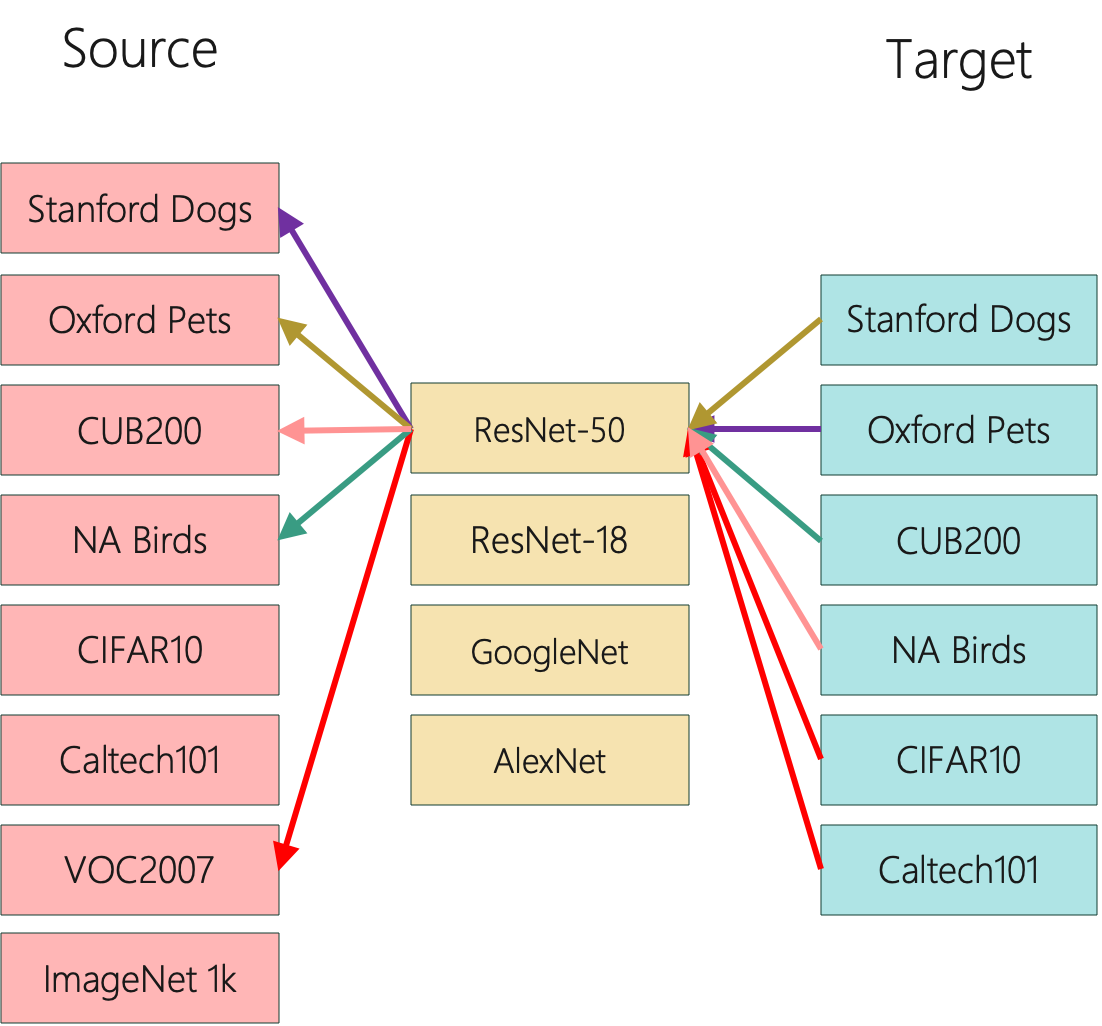}
    \caption{\textbf{\textsc{Kite} selects the pre-trainded model based on the semantics of the source dataset and the target dataset.}  }
    \label{fig: KITE_selections}
\end{figure}

In Figure \ref{fig: KITE_selections}, we show the selection of \textsc{Kite} for each target dataset. It can be observed that \textsc{Kite} selects the pre-trained model based on the semantics of the source dataset and the target dataset. The selections are also well matched to human intuition. This also indicates that transferability estimation is particularly useful when the source of the pre-trained model is unknown or the semantic similarity between the source and the target is hard to quantify.

\section{Target Alignment vs. Random Alignment}
\label{sec:ta_vs_ra}
\begin{figure*}[!t]
   \centering
 \setlength{\tabcolsep}{0pt}
%\begin{tabular}{@{\hspace{-8pt}}ccc|ccc@{}}
\begin{tabular}{ccc}
\includegraphics[width=0.3\textwidth]{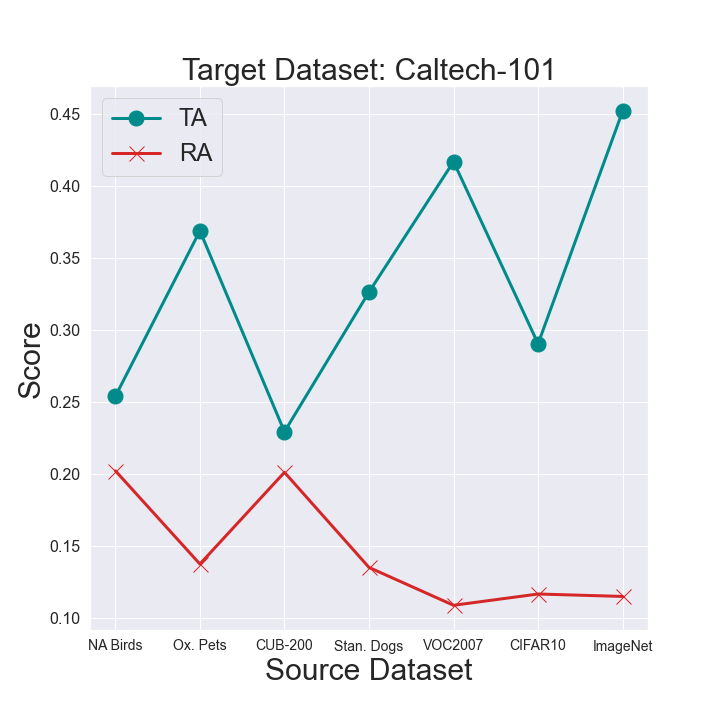}  &
\includegraphics[width=0.3\textwidth]{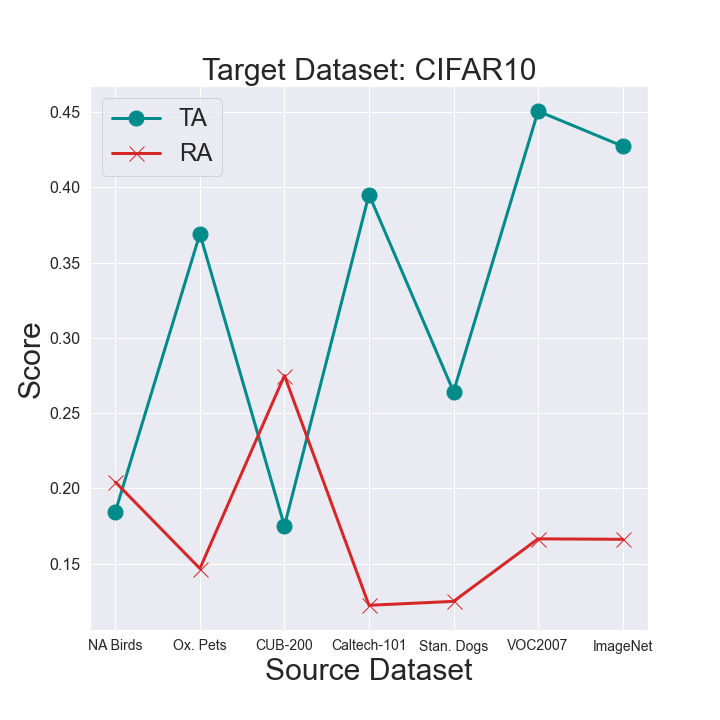}  &
\includegraphics[width=0.3\textwidth]{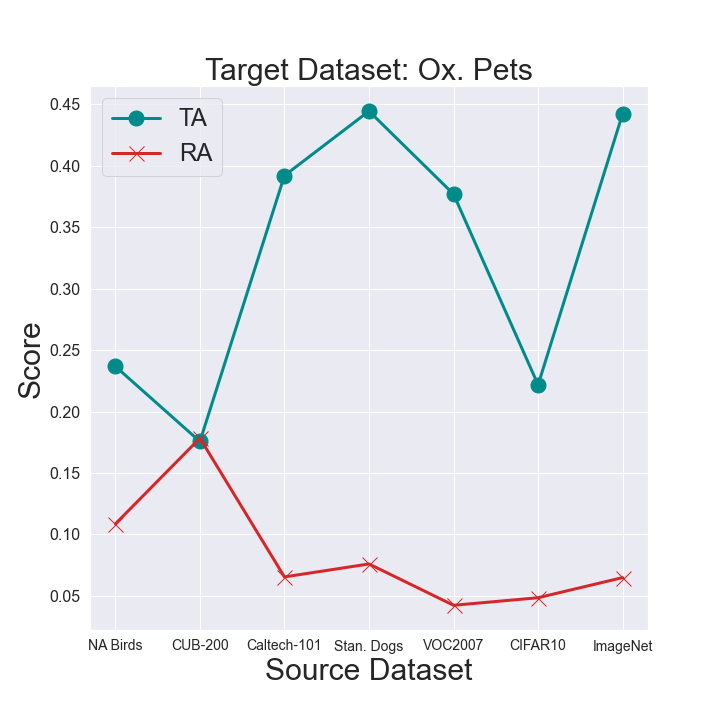}   \\
\includegraphics[width=0.3\textwidth]{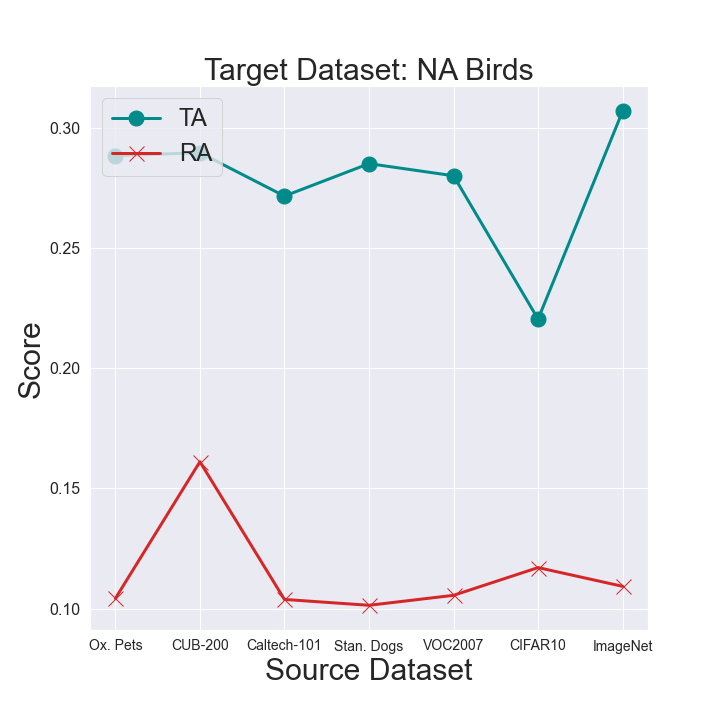}  &
\includegraphics[width=0.3\textwidth]{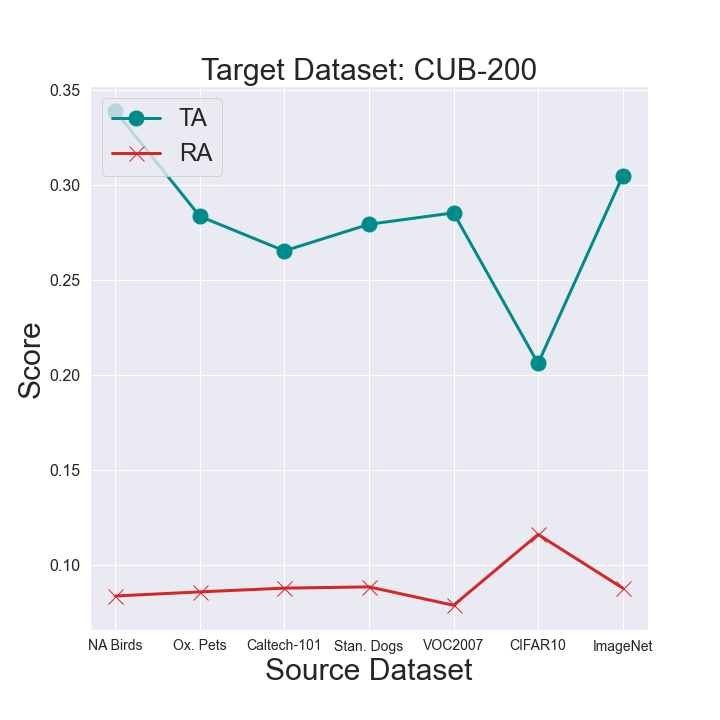}  &
\includegraphics[width=0.3\textwidth]{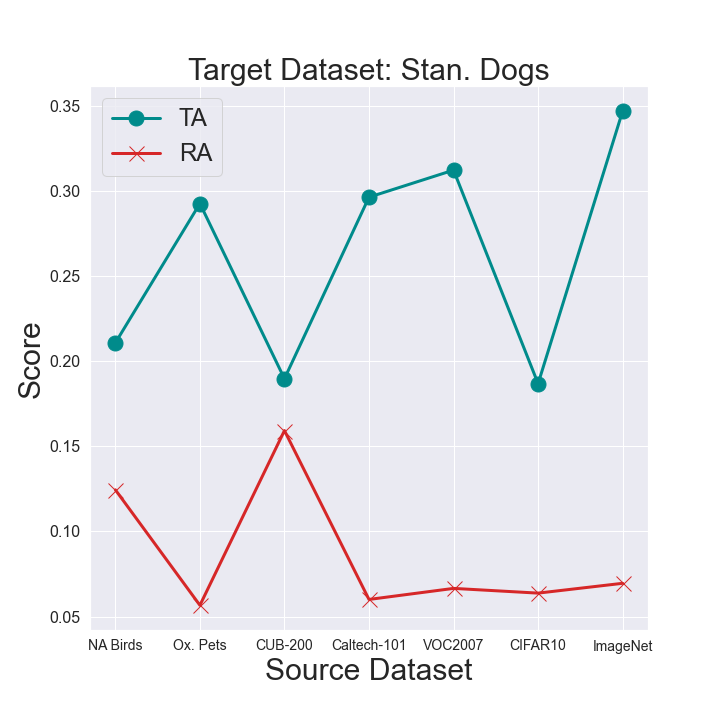}  
 \\
\end{tabular}
    \caption{The TA scores for coarse-grained datasets are higher than fine-grained datasets. Also, TA and RA have a negative correlation. }
    \label{fig:ra_ta}
\end{figure*}

Figure \ref{fig:ra_ta} shows the TA and RA scores of the features extracted by a ResNet-18 pre-trained on different source datasets on the target datasets. There are several interesting observations from the results. The first observation is the TA scores are higher for coarse-grained classification tasks (Caltech-101, CIFAR10 and Oxford Pets) compared with fine-grained classification tasks (NA Birds, CUB-200 and Stanford Dogs). \emph{We use the highest score each target dataset can achieve for ranking the separability}. This indicates that the features of coarse-grained classification tasks are inherently easier to separate. The second observation is that the TA scores and RA scores are negatively correlated. Intuitively, this means better separability indeed indicates less similarity to random features. Still, TA and RA have different implications on the properties of the feature space and different effects for transferability estimation based on the target task.

\begin{figure*}[!t]
   \centering
 \setlength{\tabcolsep}{0pt}
%\begin{tabular}{@{\hspace{-8pt}}ccc|ccc@{}}
\begin{tabular}{ccc}
\includegraphics[width=0.3\textwidth]{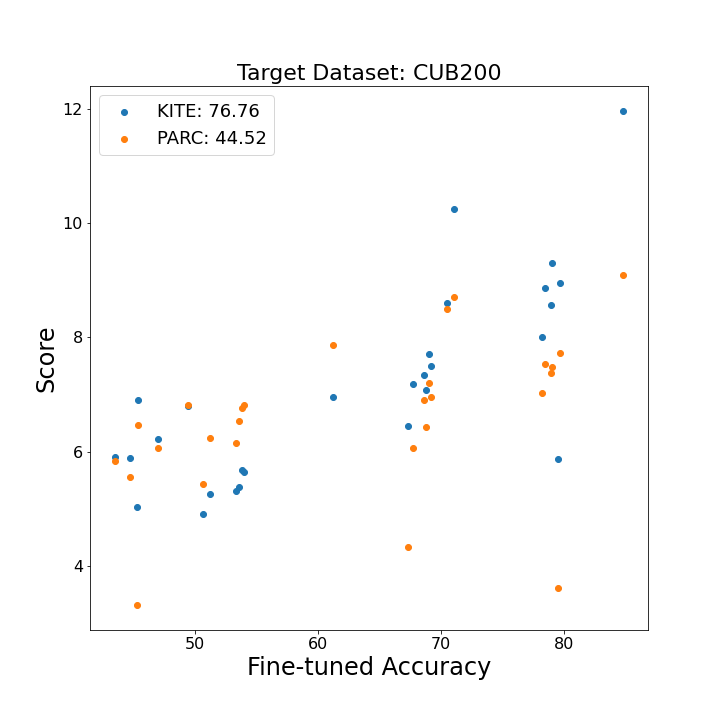} &
\includegraphics[width=0.3\textwidth]{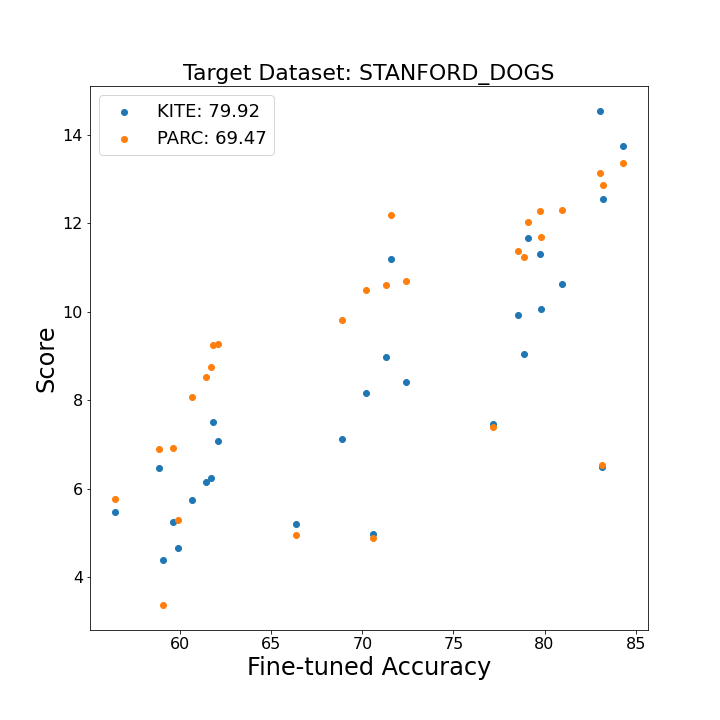}  &
\includegraphics[width=0.3\textwidth]{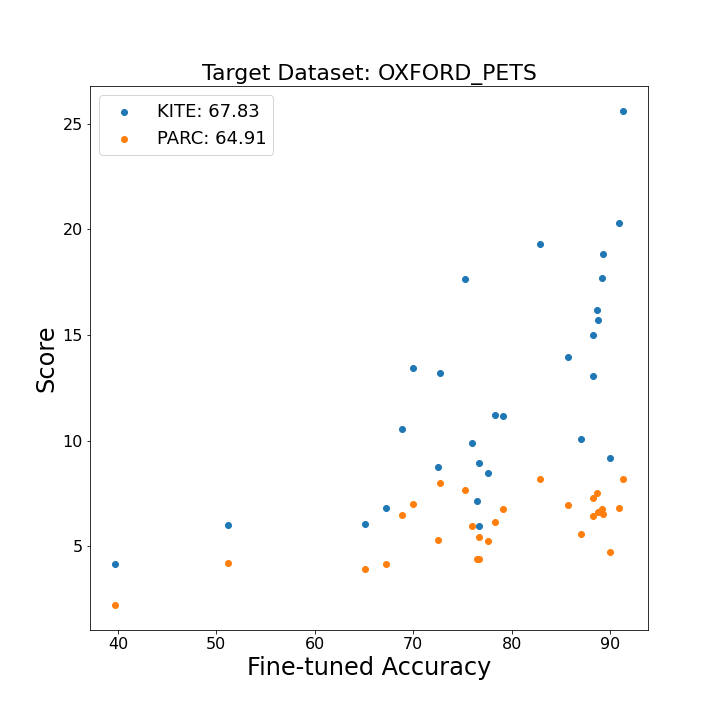}  \\
\includegraphics[width=0.3\textwidth]{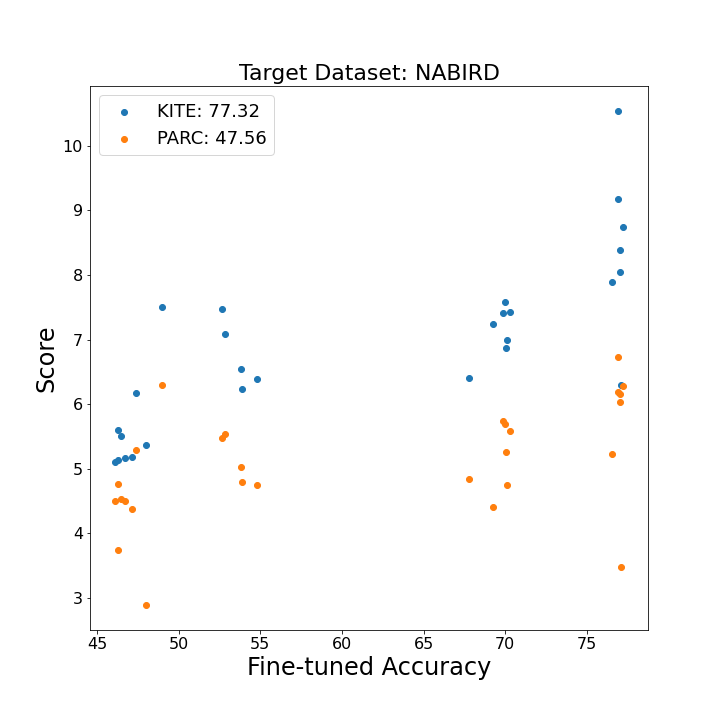} &  
\includegraphics[width=0.3\textwidth]{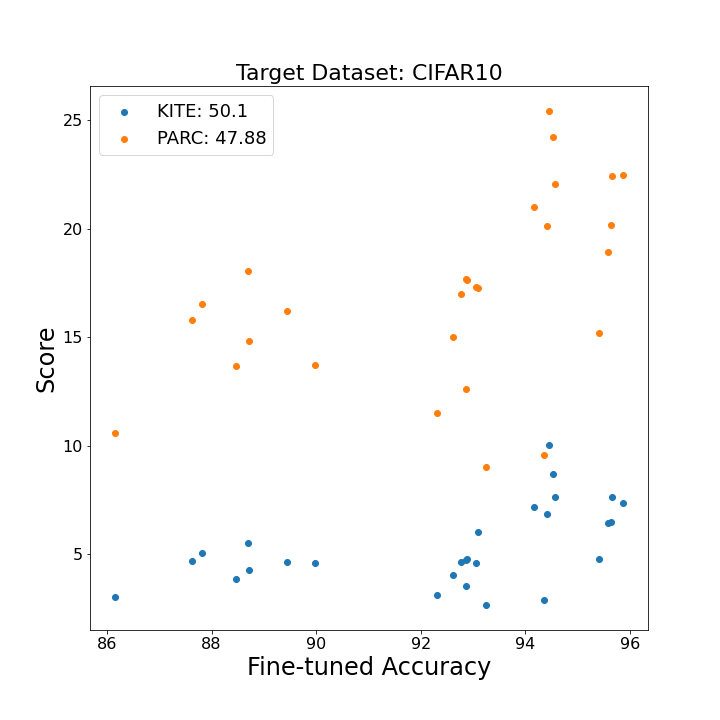} & 
\includegraphics[width=0.3\textwidth]{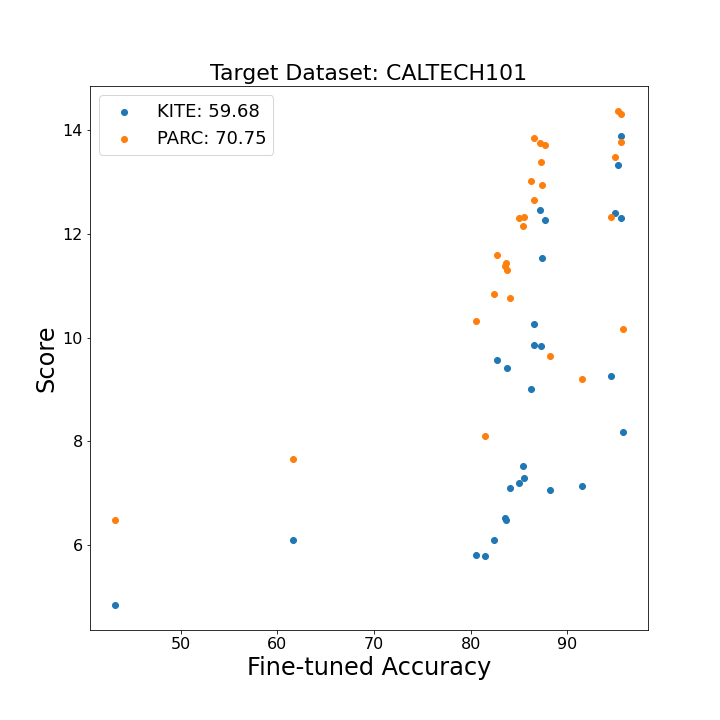} 
 \\
\end{tabular}
    \caption{\textsc{Kite} produces scores which are better correlated with the fine-tuned accuracy than PARC. For each target dataset, we compute the Pearson correlation of the computed scores with the fine-tuned accuracies obtained from all the models.}
    \label{fig:scatter}
\end{figure*}

\section{Results for Each Target Dataset} 
\label{sec:per_class}
Table \ref{tab:per_dataset_results} summarizes the  
results of the methods on each dataset. Figure \ref{fig:scatter} shows the scatter plot of the scores computed by \textsc{Kite} and PARC \cite{bolya2021scalable} versus the fine-tuned accuracies for each target dataset. Clearly, except on Caltech101, \textsc{Kite} leads to better correlation compared with PARC.

\begin{table*}[h]
    \centering
\scalebox{0.7}{\begin{tabular}{c<{\ }|r@{ $\pm$ }l<{\ }|r@{ $\pm$ }l<{\ }|r@{ $\pm$ }l<{\ }|r@{ $\pm$ }l<{\ }|r@{ $\pm$ }l<{\ }|r@{ $\pm$ }l<{\ }|r@{ $\pm$ }l<{\ }r@{ $\pm$ }l<{\ }r@{ $\pm$ }l<{\ }r@{ $\pm$ }l<{\ }r@{ $\pm$ }l<{\ }r@{ $\pm$ }l<{\ }r@{ $\pm$ }l}
    \toprule
        \multicolumn{1}{c}{\textbf{Method}}   & \multicolumn{2}{c}{Stan. Dogs} &  \multicolumn{2}{c}{Ox. Pets} &  \multicolumn{2}{c}{CUB 200} & \multicolumn{2}{c}{NA Birds} &  \multicolumn{2}{c}{CIFAR 10} &  \multicolumn{2}{c}{Caltech 101} & \multicolumn{2}{c}{\textbf{Mean PC (\%) $\uparrow$}} \\
       \midrule
        NCE \cite{tran2019transferability}   & -2.46 & 2.61 & 36.16 &  4.57 & 9.97 &  2.31 & 24.66 &  11.48  &75.20 &  4.32 & -14.37 &  1.24& 2.21 &  0.52 \\
        LEEP  \cite{nguyen2020leep}  &  30.68 & 0.34  & 39.37 & 1.98  & -1.52 & 1.14 & -17.37 & 0.94 &   76.91 & 1.62&-21.19 & 0.50 & 10.83 & 0.13 \\
        
        LogME \cite{you2021logme} & 81.12 & 0.51 & 72.09 & 1.97 & 42.75 & 0.31& 47.52 & 0.12 & 47.40 & 2.57 &65.58 & 1.15 & 55.30 & 0.41\\
        \midrule
        H-Score \cite{bao2019information} &  65.98 & 2.43 & 71.61 & 1.89 &75.84 & 3.83 &65.62 & 1.33  &46.45 & 2.39 &64.23 & 0.95 & 55.66 & 0.54\\
        
        RSA \cite{dwivedi2019representation} & -46.58 & 0.83 & -1.82 & 1.73 &-42.15 & 2.23 & -58.77 & 1.82 & 19.35 & 2.88& -23.00 & 0.44 & 5.37 & 0.57 \\    
        
        DDS \cite{dwivedi2020duality} & -38.83 & 0.72 & 11.96 & 2.33 & -40.14 & 1.94 & -59.29 & 2.12 & 22.83 & 2.11 &-17.83 & 0.79 & 10.58 & 0.32 \\
        
        PARC  \cite{bolya2021scalable} & 67.11& 1.94 & 47.29 &  18.43 & 73.44 & 3.49 &74.83 & 1.07  & 44.31 & 4.20 &59.22 & 1.26 & 59.15 & 1.17 \\
        
        GBC \cite{pandy2022transferability}  & 43.69 &1.41  & 76.48 &  0.63 & 30.78 &2.10 & 38.05 &2.29   & 38.86 & 2.52 & 80.76 &3.65  & 51.44  & 0.83 \\
        
            \midrule
          Logistic \cite{bolya2021scalable}&  70.19 &1.54& 76.50 & 6.71 & 73.54 & 8.86 & 68.80 & 10.95 & 40.00 & 11.45 &59.19 & 1.93 & 58.31 & 2.39\\
          
    1-NN CV \cite{bolya2021scalable}&61.29 & 1.91  & 81.34 & 1.86 & 71.33 & 5.22 & 72.89 & 6.77 & 58.13 & 1.42 &63.12 & 1.17 & 60.68 & 1.84\\
    
    5-NN CV \cite{bolya2021scalable}& 70.25 & 1.35 & 79.47 & 1.89 &71.69 & 7.92 &71.49 & 4.99 & 50.39 & 3.65&62.26 & 2.78 &59.72 & 1.75\\
    \midrule 
 \textsc{Kite}  &  73.86 & 1.72 &  72.21 & 3.33  &   80.29 & 0.87 &   86.48 & 2.23 &  39.52 & 2.15 &64.84 & 1.88 & 67.90 & 0.70 \\
\bottomrule
\end{tabular}}
    \caption{The results of all the methods for each target dataset.} 
    \label{tab:per_dataset_results}
\end{table*}

\begin{figure*}[!t]
   \centering
 \setlength{\tabcolsep}{0pt}
%\begin{tabular}{@{\hspace{-8pt}}ccc|ccc@{}}
\begin{tabular}{cccc}
\includegraphics[width=0.24\textwidth]{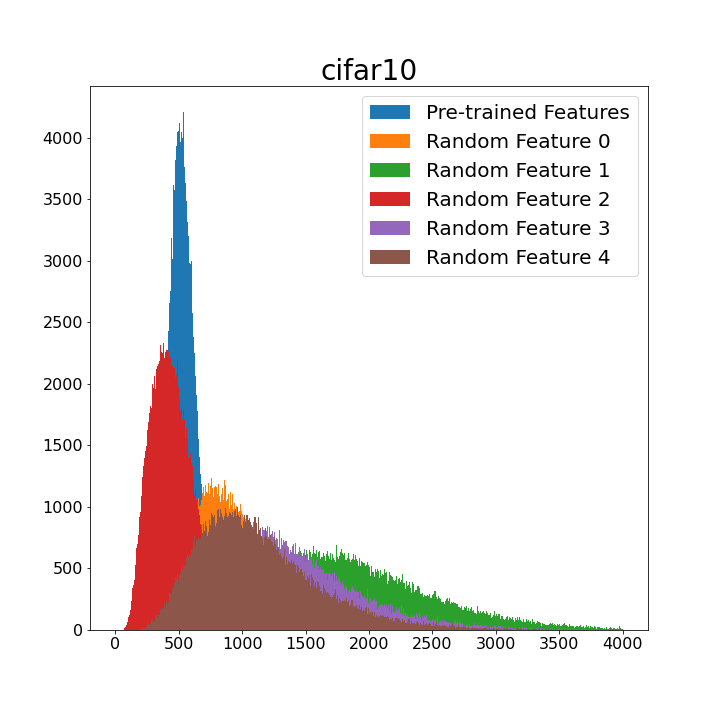}  &
\includegraphics[width=0.24\textwidth]{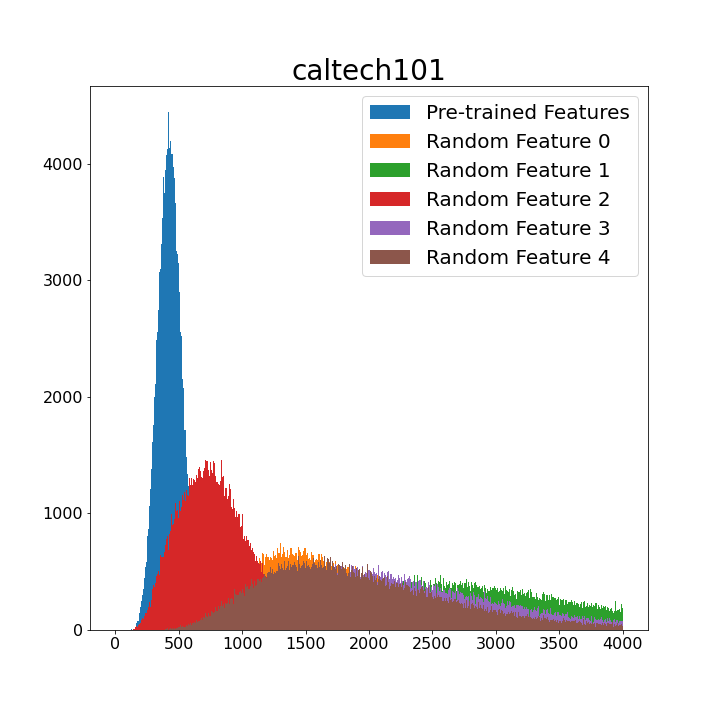}  &
\includegraphics[width=0.24\textwidth]{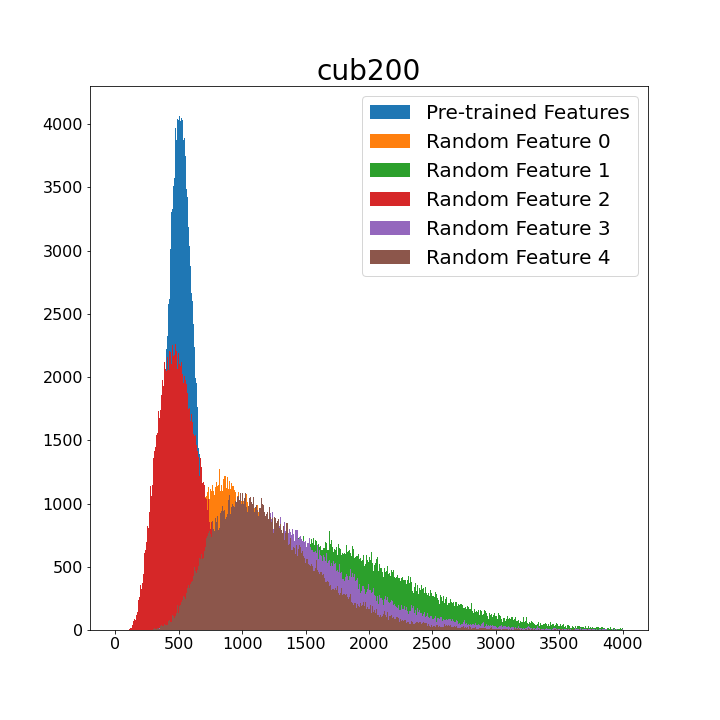}
 &
\includegraphics[width=0.24\textwidth]{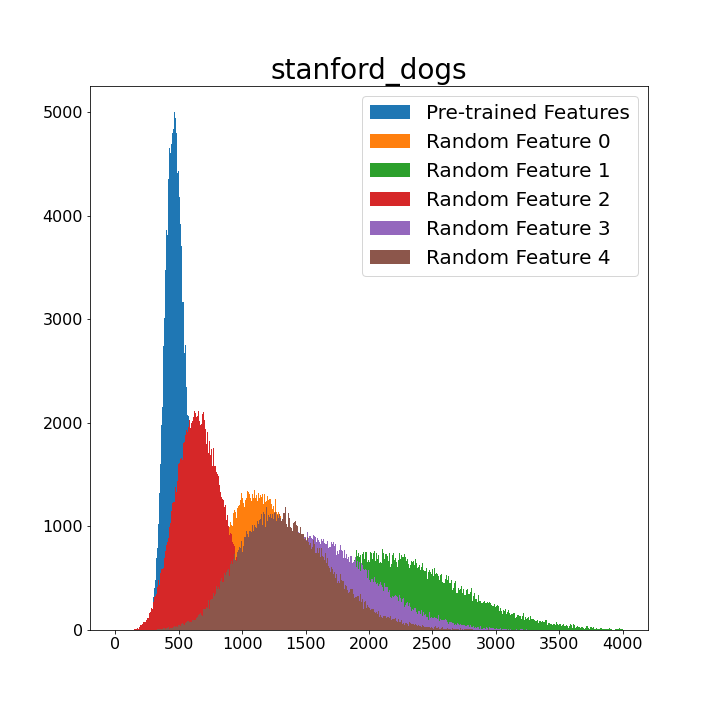}
\end{tabular}
    \caption{The similarity values of the pre-trained kernel matrix $\mathbf{K}_s$ are more concentrated compared with the values of the random kernel matrices. Intuitively, since the pre-trained features already capture data similarity, the number of small similarity values will be dominant, i.e., the similarity values are more concentrated. On the other hand, the similarity values of the random kernel matrix are more scattered.  We use pre-trained ResNet18 on ImageNet as the source model and use CIFAR10, Caltech101, CUB200 and Stanford Dogs as the target dataaset to generate the features. The random kernel matrices are generated with different random seeds (0 to 4). }
    \label{fig:comparision_random}
\end{figure*}

\begin{figure}
    \centering
    \includegraphics[width=0.45\textwidth]{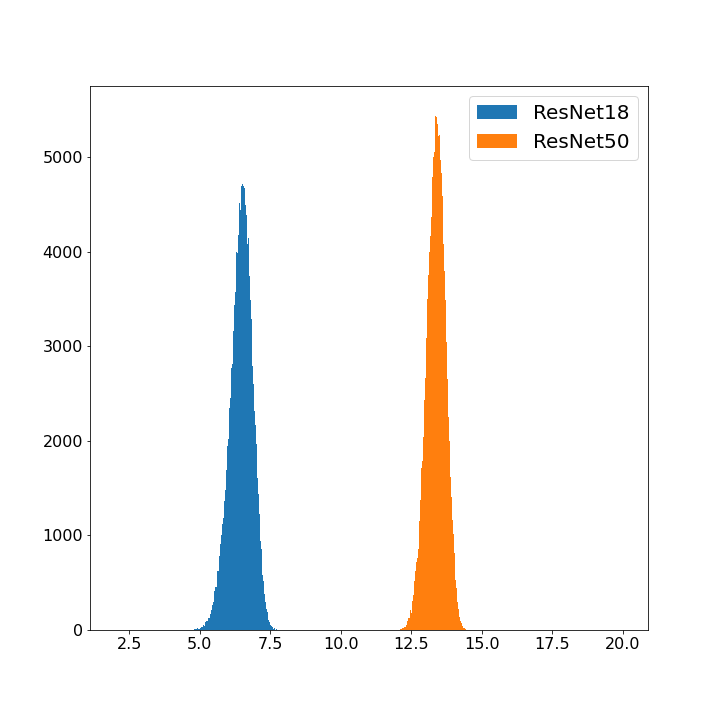}
    \caption{Distribution of the random similarity values are different for different architectures. The similarity values are on a log scale. We generate $\mathbf{K}_{random}$ on CUB200 by using different architectures which are pre-trained on ImageNet.  }
    \label{fig:random_hist}
\end{figure}

\section{More Discussion on the Random Kernel Matrix}
\label{sec:random_kernel}

The study of the random kernel matrix $\mathbf{K}_{random}$ is of great theoretical interest \cite{pennington2017nonlinear,louart2018random,liao2020random,chen2022concentration}. To study the limiting behavior of $\mathbf{K}_{random}$ is difficult due to the composition of multiple layers \cite{louart2018random,liao2020random}. In this paper, we view $\mathbf{K}_{random}$ as a poor approximation of true data similarity. If we vectorize the random kernel matrix as $\vect (\mathbf{K}_{random})$, it can be viewed as a high-dimensional random vector which has an unknown distribution. The values can be regarded as random similarity values. Nevertheless, we can approximate the distribution by generating $\vect (\mathbf{K}_{random})$ using different random seeds. Moreover, we found that the concentration of $\vect (\mathbf{K}_{random})$ is affected by the target dataset and the architecture.

Figure \ref{fig:comparision_random} shows that the distribution of the values in the pre-trained feature kernel matrix are different from that of the random kernel matrix. Thus, the dissimilarity to the random features can be an indicator for the transferability of the pre-trained models.

Figure \ref{fig:random_hist} shows that distribution of the random similarity values are different for different architectures. We generate $\mathbf{K}_{random}$ on CUB200 by using different architectures which are pre-trained on ImageNet. This shows that architecture-specific random features are important for characterizing the usefulness of the pre-trained features.

\section{Discussion on Limitations and Extensions}
\label{sec:limitation}

This paper addresses an important problem in machine learning, and we believe the proposed method should not raise any ethical considerations. We discuss limitations and possible extensions below,

\noindent \textbf{Computational Resources.} In general, transferability estimation methods prefer to select deeper model. This is a desirable property if the computational resource is not a bottleneck. In practice, the users may not have enough resources. Thus, it would be interesting to take the actual resources the users have into account for transferability estimation.

\noindent \textbf{Downstream Tasks.} Currently, we only consider classification as the downstream task. It would be interesting to extend \textsc{Kite} for downstream tasks such as instance segmentation, semantic segmentation and object detection. To accurately estimate the fine-tuning performance of pre-trained models would greatly improve the performance of these tasks.
%%%%%%%%% REFERENCES
{\small
\bibliographystyle{ieee_fullname}
\bibliography{egbib}

\begin{thebibliography}{10}\itemsep=-1pt

\bibitem{abdi2010principal}
Herv{\'e} Abdi and Lynne~J Williams.
\newblock Principal component analysis.
\newblock {\em Wiley interdisciplinary reviews: computational statistics}, 2(4):433--459, 2010.

\bibitem{badrinarayanan2017segnet}
Vijay Badrinarayanan, Alex Kendall, and Roberto Cipolla.
\newblock Segnet: A deep convolutional encoder-decoder architecture for image segmentation.
\newblock {\em IEEE transactions on pattern analysis and machine intelligence}, 39(12):2481--2495, 2017.

\bibitem{bao2019information}
Yajie Bao, Yang Li, Shao-Lun Huang, Lin Zhang, Lizhong Zheng, Amir Zamir, and Leonidas Guibas.
\newblock An information-theoretic approach to transferability in task transfer learning.
\newblock In {\em 2019 IEEE International Conference on Image Processing (ICIP)}, pages 2309--2313. IEEE, 2019.

\bibitem{bengio2012deep}
Yoshua Bengio.
\newblock Deep learning of representations for unsupervised and transfer learning.
\newblock In {\em Proceedings of ICML workshop on unsupervised and transfer learning}, pages 17--36. JMLR Workshop and Conference Proceedings, 2012.

\bibitem{bolya2021scalable}
Daniel Bolya, Rohit Mittapalli, and Judy Hoffman.
\newblock Scalable diverse model selection for accessible transfer learning.
\newblock {\em Advances in Neural Information Processing Systems}, 34, 2021.

\bibitem{chen2020simple}
Ting Chen, Simon Kornblith, Mohammad Norouzi, and Geoffrey Hinton.
\newblock A simple framework for contrastive learning of visual representations.
\newblock In {\em International conference on machine learning}, pages 1597--1607. PMLR, 2020.

\bibitem{chen2022concentration}
Zhijun Chen, Hayden Schaeffer, and Rachel Ward.
\newblock Concentration of random feature matrices in high-dimensions.
\newblock {\em arXiv preprint arXiv:2204.06935}, 2022.

\bibitem{cortes2012algorithms}
Corinna Cortes, Mehryar Mohri, and Afshin Rostamizadeh.
\newblock Algorithms for learning kernels based on centered alignment.
\newblock {\em The Journal of Machine Learning Research}, 13:795--828, 2012.

\bibitem{cristianini2001kernel}
Nello Cristianini, John Shawe-Taylor, Andre Elisseeff, and Jaz Kandola.
\newblock On kernel-target alignment.
\newblock {\em Advances in neural information processing systems}, 14, 2001.

\bibitem{deng2009imagenet}
Jia Deng, Wei Dong, Richard Socher, Li-Jia Li, Kai Li, and Li Fei-Fei.
\newblock Imagenet: A large-scale hierarchical image database.
\newblock In {\em 2009 IEEE conference on computer vision and pattern recognition}, pages 248--255. Ieee, 2009.

\bibitem{donahue2014decaf}
Jeff Donahue, Yangqing Jia, Oriol Vinyals, Judy Hoffman, Ning Zhang, Eric Tzeng, and Trevor Darrell.
\newblock Decaf: A deep convolutional activation feature for generic visual recognition.
\newblock In {\em International conference on machine learning}, pages 647--655. PMLR, 2014.

\bibitem{dwivedi2020duality}
Kshitij Dwivedi, Jiahui Huang, Radoslaw~Martin Cichy, and Gemma Roig.
\newblock Duality diagram similarity: a generic framework for initialization selection in task transfer learning.
\newblock In {\em European Conference on Computer Vision}, pages 497--513. Springer, 2020.

\bibitem{dwivedi2019representation}
Kshitij Dwivedi and Gemma Roig.
\newblock Representation similarity analysis for efficient task taxonomy \& transfer learning.
\newblock In {\em Proceedings of the IEEE/CVF Conference on Computer Vision and Pattern Recognition}, pages 12387--12396, 2019.

\bibitem{everingham2010pascal}
Mark Everingham, Luc Van~Gool, Christopher~KI Williams, John Winn, and Andrew Zisserman.
\newblock The pascal visual object classes (voc) challenge.
\newblock {\em International journal of computer vision}, 88(2):303--338, 2010.

\bibitem{fei2006one}
Li Fei-Fei, Rob Fergus, and Pietro Perona.
\newblock One-shot learning of object categories.
\newblock {\em IEEE transactions on pattern analysis and machine intelligence}, 28(4):594--611, 2006.

\bibitem{freedman2007statistics}
David Freedman, Robert Pisani, and Roger Purves.
\newblock Statistics (international student edition).
\newblock {\em Pisani, R. Purves, 4th edn. WW Norton \& Company, New York}, 2007.

\bibitem{ge2017borrowing}
Weifeng Ge and Yizhou Yu.
\newblock Borrowing treasures from the wealthy: Deep transfer learning through selective joint fine-tuning.
\newblock In {\em Proceedings of the IEEE conference on computer vision and pattern recognition}, pages 1086--1095, 2017.

\bibitem{girshick2014rich}
Ross Girshick, Jeff Donahue, Trevor Darrell, and Jitendra Malik.
\newblock Rich feature hierarchies for accurate object detection and semantic segmentation.
\newblock In {\em Proceedings of the IEEE conference on computer vision and pattern recognition}, pages 580--587, 2014.

\bibitem{glorot2010understanding}
Xavier Glorot and Yoshua Bengio.
\newblock Understanding the difficulty of training deep feedforward neural networks.
\newblock In {\em Proceedings of the thirteenth international conference on artificial intelligence and statistics}, pages 249--256. JMLR Workshop and Conference Proceedings, 2010.

\bibitem{gretton2005measuring}
Arthur Gretton, Olivier Bousquet, Alex Smola, and Bernhard Sch{\"o}lkopf.
\newblock Measuring statistical dependence with hilbert-schmidt norms.
\newblock In {\em International conference on algorithmic learning theory}, pages 63--77. Springer, 2005.

\bibitem{grill2020bootstrap}
Jean-Bastien Grill, Florian Strub, Florent Altch{\'e}, Corentin Tallec, Pierre Richemond, Elena Buchatskaya, Carl Doersch, Bernardo Avila~Pires, Zhaohan Guo, Mohammad Gheshlaghi~Azar, et~al.
\newblock Bootstrap your own latent-a new approach to self-supervised learning.
\newblock {\em Advances in Neural Information Processing Systems}, 33:21271--21284, 2020.

\bibitem{guo2020adafilter}
Yunhui Guo, Yandong Li, Liqiang Wang, and Tajana Rosing.
\newblock Adafilter: Adaptive filter fine-tuning for deep transfer learning.
\newblock In {\em Proceedings of the AAAI Conference on Artificial Intelligence}, volume~34, pages 4060--4066, 2020.

\bibitem{guo2019spottune}
Yunhui Guo, Honghui Shi, Abhishek Kumar, Kristen Grauman, Tajana Rosing, and Rogerio Feris.
\newblock Spottune: transfer learning through adaptive fine-tuning.
\newblock In {\em Proceedings of the IEEE/CVF conference on computer vision and pattern recognition}, pages 4805--4814, 2019.

\bibitem{he2020momentum}
Kaiming He, Haoqi Fan, Yuxin Wu, Saining Xie, and Ross Girshick.
\newblock Momentum contrast for unsupervised visual representation learning.
\newblock In {\em Proceedings of the IEEE/CVF conference on computer vision and pattern recognition}, pages 9729--9738, 2020.

\bibitem{he2019rethinking}
Kaiming He, Ross Girshick, and Piotr Doll{\'a}r.
\newblock Rethinking imagenet pre-training.
\newblock In {\em Proceedings of the IEEE/CVF International Conference on Computer Vision}, pages 4918--4927, 2019.

\bibitem{he2015delving}
Kaiming He, Xiangyu Zhang, Shaoqing Ren, and Jian Sun.
\newblock Delving deep into rectifiers: Surpassing human-level performance on imagenet classification.
\newblock In {\em Proceedings of the IEEE international conference on computer vision}, pages 1026--1034, 2015.

\bibitem{he2016deep}
Kaiming He, Xiangyu Zhang, Shaoqing Ren, and Jian Sun.
\newblock Deep residual learning for image recognition.
\newblock In {\em Proceedings of the IEEE conference on computer vision and pattern recognition}, pages 770--778, 2016.

\bibitem{hinton2007recognize}
Geoffrey~E Hinton.
\newblock To recognize shapes, first learn to generate images.
\newblock {\em Progress in brain research}, 165:535--547, 2007.

\bibitem{khosla2011novel}
Aditya Khosla, Nityananda Jayadevaprakash, Bangpeng Yao, and Fei-Fei Li.
\newblock Novel dataset for fine-grained image categorization: Stanford dogs.
\newblock In {\em Proc. CVPR Workshop on Fine-Grained Visual Categorization (FGVC)}, volume~2. Citeseer, 2011.

\bibitem{knuth2015bayesian}
Kevin~H Knuth, Michael Habeck, Nabin~K Malakar, Asim~M Mubeen, and Ben Placek.
\newblock Bayesian evidence and model selection.
\newblock {\em Digital Signal Processing}, 47:50--67, 2015.

\bibitem{kornblith2021better}
Simon Kornblith, Ting Chen, Honglak Lee, and Mohammad Norouzi.
\newblock Why do better loss functions lead to less transferable features?
\newblock {\em Advances in Neural Information Processing Systems}, 34, 2021.

\bibitem{kornblith2019better}
Simon Kornblith, Jonathon Shlens, and Quoc~V Le.
\newblock Do better imagenet models transfer better?
\newblock In {\em Proceedings of the IEEE/CVF conference on computer vision and pattern recognition}, pages 2661--2671, 2019.

\bibitem{krizhevsky2009learning}
Alex Krizhevsky, Geoffrey Hinton, et~al.
\newblock Learning multiple layers of features from tiny images.
\newblock 2009.

\bibitem{krizhevsky2012imagenet}
Alex Krizhevsky, Ilya Sutskever, and Geoffrey~E Hinton.
\newblock Imagenet classification with deep convolutional neural networks.
\newblock {\em Advances in neural information processing systems}, 25, 2012.

\bibitem{liao2020random}
Zhenyu Liao, Romain Couillet, and Michael~W Mahoney.
\newblock A random matrix analysis of random fourier features: beyond the gaussian kernel, a precise phase transition, and the corresponding double descent.
\newblock {\em Advances in Neural Information Processing Systems}, 33:13939--13950, 2020.

\bibitem{long2015fully}
Jonathan Long, Evan Shelhamer, and Trevor Darrell.
\newblock Fully convolutional networks for semantic segmentation.
\newblock In {\em Proceedings of the IEEE conference on computer vision and pattern recognition}, pages 3431--3440, 2015.

\bibitem{louart2018random}
Cosme Louart, Zhenyu Liao, and Romain Couillet.
\newblock A random matrix approach to neural networks.
\newblock {\em The Annals of Applied Probability}, 28(2):1190--1248, 2018.

\bibitem{neyshabur2020being}
Behnam Neyshabur, Hanie Sedghi, and Chiyuan Zhang.
\newblock What is being transferred in transfer learning?
\newblock {\em Advances in neural information processing systems}, 33:512--523, 2020.

\bibitem{nguyen2020leep}
Cuong Nguyen, Tal Hassner, Matthias Seeger, and Cedric Archambeau.
\newblock Leep: A new measure to evaluate transferability of learned representations.
\newblock In {\em International Conference on Machine Learning}, pages 7294--7305. PMLR, 2020.

\bibitem{pandy2022transferability}
Michal P{\'a}ndy, Andrea Agostinelli, Jasper Uijlings, Vittorio Ferrari, and Thomas Mensink.
\newblock Transferability estimation using bhattacharyya class separability.
\newblock In {\em Proceedings of the IEEE/CVF Conference on Computer Vision and Pattern Recognition}, pages 9172--9182, 2022.

\bibitem{parkhi2012cats}
Omkar~M Parkhi, Andrea Vedaldi, Andrew Zisserman, and CV Jawahar.
\newblock Cats and dogs.
\newblock In {\em 2012 IEEE conference on computer vision and pattern recognition}, pages 3498--3505. IEEE, 2012.

\bibitem{pennington2017nonlinear}
Jeffrey Pennington and Pratik Worah.
\newblock Nonlinear random matrix theory for deep learning.
\newblock {\em Advances in neural information processing systems}, 30, 2017.

\bibitem{raghu2019transfusion}
Maithra Raghu, Chiyuan Zhang, Jon Kleinberg, and Samy Bengio.
\newblock Transfusion: Understanding transfer learning for medical imaging.
\newblock {\em Advances in neural information processing systems}, 32, 2019.

\bibitem{rahimi2007random}
Ali Rahimi and Benjamin Recht.
\newblock Random features for large-scale kernel machines.
\newblock {\em Advances in neural information processing systems}, 20, 2007.

\bibitem{ren2015faster}
Shaoqing Ren, Kaiming He, Ross Girshick, and Jian Sun.
\newblock Faster r-cnn: Towards real-time object detection with region proposal networks.
\newblock {\em Advances in neural information processing systems}, 28, 2015.

\bibitem{sharif2014cnn}
Ali Sharif~Razavian, Hossein Azizpour, Josephine Sullivan, and Stefan Carlsson.
\newblock Cnn features off-the-shelf: an astounding baseline for recognition.
\newblock In {\em Proceedings of the IEEE conference on computer vision and pattern recognition workshops}, pages 806--813, 2014.

\bibitem{smola1998learning}
Alex~J Smola and Bernhard Sch{\"o}lkopf.
\newblock {\em Learning with kernels}, volume~4.
\newblock Citeseer, 1998.

\bibitem{szegedy2015going}
Christian Szegedy, Wei Liu, Yangqing Jia, Pierre Sermanet, Scott Reed, Dragomir Anguelov, Dumitru Erhan, Vincent Vanhoucke, and Andrew Rabinovich.
\newblock Going deeper with convolutions.
\newblock In {\em Proceedings of the IEEE conference on computer vision and pattern recognition}, pages 1--9, 2015.

\bibitem{tran2019transferability}
Anh~T Tran, Cuong~V Nguyen, and Tal Hassner.
\newblock Transferability and hardness of supervised classification tasks.
\newblock In {\em Proceedings of the IEEE/CVF International Conference on Computer Vision}, pages 1395--1405, 2019.

\bibitem{van2008visualizing}
Laurens Van~der Maaten and Geoffrey Hinton.
\newblock Visualizing data using t-sne.
\newblock {\em Journal of machine learning research}, 9(11), 2008.

\bibitem{van2015building}
Grant Van~Horn, Steve Branson, Ryan Farrell, Scott Haber, Jessie Barry, Panos Ipeirotis, Pietro Perona, and Serge Belongie.
\newblock Building a bird recognition app and large scale dataset with citizen scientists: The fine print in fine-grained dataset collection.
\newblock In {\em Proceedings of the IEEE Conference on Computer Vision and Pattern Recognition}, pages 595--604, 2015.

\bibitem{vigna2015weighted}
Sebastiano Vigna.
\newblock A weighted correlation index for rankings with ties.
\newblock In {\em Proceedings of the 24th international conference on World Wide Web}, pages 1166--1176, 2015.

\bibitem{wah2011caltech}
Catherine Wah, Steve Branson, Peter Welinder, Pietro Perona, and Serge Belongie.
\newblock The caltech-ucsd birds-200-2011 dataset.
\newblock 2011.

\bibitem{xuhong2018explicit}
LI Xuhong, Yves Grandvalet, and Franck Davoine.
\newblock Explicit inductive bias for transfer learning with convolutional networks.
\newblock In {\em International Conference on Machine Learning}, pages 2825--2834. PMLR, 2018.

\bibitem{yosinski2014transferable}
Jason Yosinski, Jeff Clune, Yoshua Bengio, and Hod Lipson.
\newblock How transferable are features in deep neural networks?
\newblock {\em Advances in neural information processing systems}, 27, 2014.

\bibitem{you2021logme}
Kaichao You, Yong Liu, Jianmin Wang, and Mingsheng Long.
\newblock Logme: Practical assessment of pre-trained models for transfer learning.
\newblock In {\em International Conference on Machine Learning}, pages 12133--12143. PMLR, 2021.

\bibitem{zamir2018taskonomy}
Amir~R Zamir, Alexander Sax, William Shen, Leonidas~J Guibas, Jitendra Malik, and Silvio Savarese.
\newblock Taskonomy: Disentangling task transfer learning.
\newblock In {\em Proceedings of the IEEE conference on computer vision and pattern recognition}, pages 3712--3722, 2018.

\end{thebibliography}
}

\end{document}